\pdfoutput=1
\documentclass[11pt]{article}
\usepackage[utf8]{inputenc} 
\usepackage[T1]{fontenc}    
\usepackage{hyperref}       
\usepackage{url}            
\usepackage{booktabs}       
\usepackage{amsfonts}       
\usepackage{nicefrac}       
\usepackage{microtype}      
\usepackage{lipsum}

\usepackage{amssymb, amsmath, latexsym, amsthm}
\usepackage{comment}
\usepackage{graphicx}
\usepackage{authblk}
\usepackage{geometry}
\usepackage{natbib}
\usepackage{bm}

\def\B{{\bm B}}
\def\b{{\bm b}}

\def\d{{\bm d}}

\def\E{{\bm E}}

\def\L{{\bm L}}
\def\f{{\bm f}}

\def\m{{\bm m}}
\def\P{{\bm P}}
\def\x{{\bm x}}
\def\y{{\bm y}}
\def\I{{\bm I}}
\def\X{{\bm X}}
\def\Y{{\bm Y}}
\def\u{{\bm u}}

\def\0{{\bm 0}}
\def\1{{\bm 1}}
\def\K{{\bm K}}

\def\thet{{\boldsymbol \theta}}
\def\XI{{\boldsymbol \xi}}
\def\Rbb{\mathbb{R}}
\def\Ebb{\mathbb{E}}

\def\mX{{\mathcal{X}}}

\def\mGP{{\mathcal{GP}}}

\newcommand{\argmin}{\mathop{\rm arg~min}\limits}
\newcommand{\argmax}{\mathop{\rm arg~max}\limits}

\makeatletter
\newcommand{\figcaption}[1]{\def\@captype{figure}\caption{#1}}
\newcommand{\tblcaption}[1]{\def\@captype{table}\caption{#1}}
\makeatother

\RequirePackage{algorithm}
\RequirePackage{algorithmic}
\usepackage{bm}

\def\<{\langle}
\def\>{\rangle}

\usepackage{color}

\title{Bayesian Active Learning 
for Structured Output Design}

\author[1]{Kota Matsui\thanks{equal contribution}\thanks{kota.matsui@riken.jp}}
\author[2]{Shunya Kusakawa$^*$}
\author[3]{Keisuke Ando}
\author[1]{Kentaro Kutsukake}
\author[3,4]{Toru Ujihara}
\author[2,1]{Ichiro Takeuchi}

\affil[1]{RIKEN AIP, Japan}
\affil[2]{Nagoya Institute of Technology, Japan}
\affil[3]{Nagoya University, Japan}
\affil[4]{AIST, Japan}

\date{}

\begin{document}

\maketitle

\begin{abstract}
  In this paper, we propose an active learning method for an inverse problem that aims to find an input that achieves a desired structured-output. The proposed method provides new acquisition functions for minimizing the error between the desired structured-output and the prediction of a Gaussian process model, by effectively incorporating the correlation between multiple outputs of the underlying multi-valued black box output functions. The effectiveness of the proposed method is verified by applying it to two synthetic shape search problem and real data. In the real data experiment, we tackle the input parameter search which achieves the desired crystal growth rate in silicon carbide (SiC) crystal growth modeling, that is a problem of materials informatics. 

\end{abstract}


\section{Introduction}
\label{intro}
Let us consider the problem of finding $\x_0 \in \mathcal{X}$ from
given $\y_0 \in \mathcal{Y}$ where $\x_0$ and $\y_0$ are corresponding input and output, i.e.,
\begin{align*}
    \y_0 = f(\x_0).
\end{align*}
These type of problems are known as {\it inverse problems}.
Inverse problems have been widely studied, especially in the field of mathematics and physics~\citep{kirsch2011introduction}.
Indeed, there are many such problems that arise in science, for example the problem of estimating the internal parameters of a simulator of physical phenomena.
In materials science, numerical calculations using simulators are widely used to predict the results of actual experiments against unknown experimental conditions. However, there are many simulators that are not accurate (in other words, the actual experimental results and the simulator's predicted results are different).
This is presumably due to the fact that the internal parameters of the simulator are not set correctly, and the estimation accuracy could be expected to be improved by limiting the range of the internal parameters using the observed actual experimental results~\citep{adachi2005homogeneous, daggupati2010solid}.
Such problems can be formulated as an inverse problem that involves finding the internal parameters that achieve a desired experimental result (often reffered to as a structural output or ``shape").
Therefore, our goal is to solve an inverse problem, the result of which is output with the desired structure or ``shape".
Since, in general, physical systems are often black-box, we approach this problem
in a data-driven way by utilizing the observable input-output pairs as data.

Furthermore, many scientific and engineering problems are accompanied by
uncertainties for various reasons such as lack of data or noisy observations.
In addition, due to realistic constraints such as the costliness of conducting experiments,
it is desirable to minimize the number of observations that need to be gathered from actual
experiments.
Active learning (or statistical experimental design, as it is also known)~\citep{settles2009active} is one of the most effective machine learning methods for such problems.
In particular, for black box functions such as those that often comprise physical systems, an active learning method (also known as Bayesian optimization)~\citep{shahriari2015taking} that introduces a Gaussian process model into the objective function and selects the next observation point by taking into account the model uncertainty, is an approach that has been widely studied.

\begin{figure}[t]
 \begin{center}
 \begin{tabular}{cc}
 \hspace{-0.9cm}
  \includegraphics[bb = 11 12 707 527, scale=0.33]{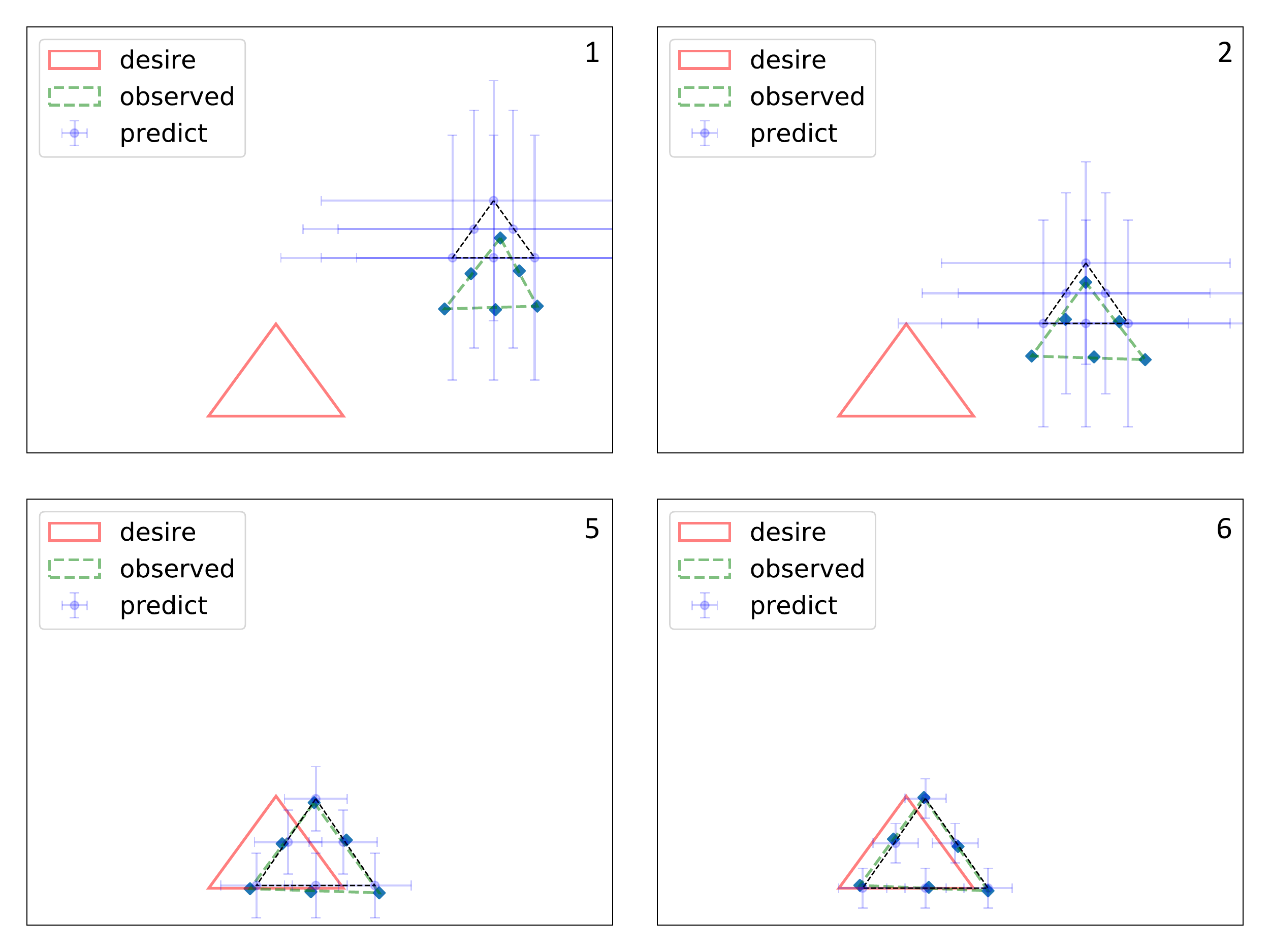} & 
  \hspace{-0.6cm}
  \includegraphics[bb = 11 12 707 527, scale=0.33]{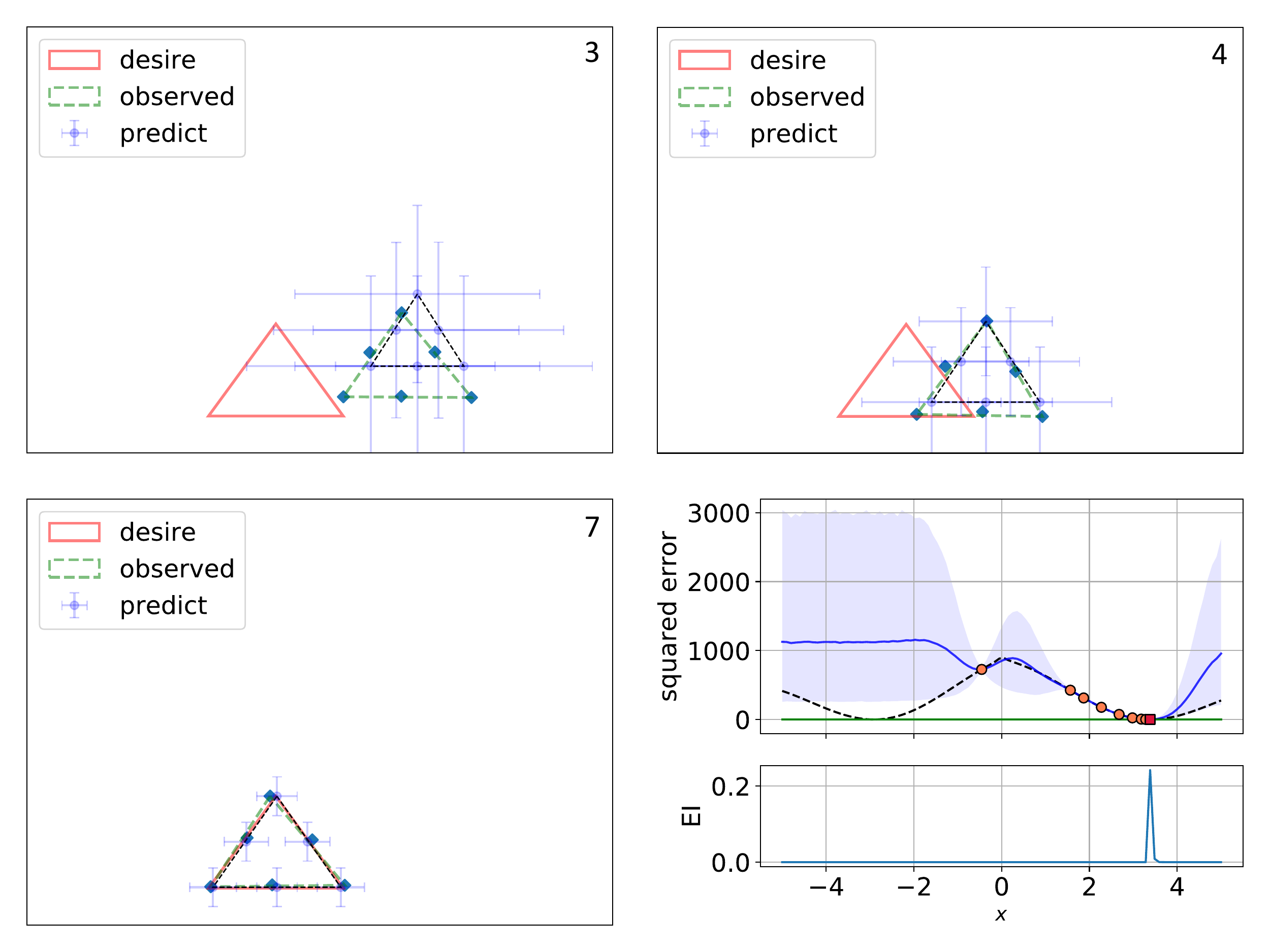}
 \end{tabular}
 \caption{
 {\footnotesize 
 Demonstration of the proposed method. 
 The search proceeds from the top left to the 
 bottom right. 
 The right end plot of the bottom row shows the final state of the model. 
 We can see that based on the model (blue triangle) for the true structural output (red triangle), the next observation data (green triangle) can be selected appropriately, and the desired output can be obtained with a small number of observations. The details are 
 explained in Section~\ref{experiments}. 
 }
 \label{fig:demo_intro}
 }
 \end{center}
\end{figure}

In this paper, we tackle the inverse problem for black-box functions which return
noisy, structured-output.
The proposed method is formulated as a simple error minimization algorithm for vector-valued functions. Nevertheless, the novel acquisition function that takes into account the structure of the output (i.e. the correlation between functions) can efficiently find the input that achieves a desired structured output.
Figure~\ref{fig:demo_intro} demonstrates the proposed method.
We can see that the red triangle (the desired structured output) can be achieved efficiently
by using the prediction of the proposed method as a guide.

\subsection*{Contributions}

Our contributions are summarized as follows. 
\begin{itemize}
    \item We formulate a statistical experimental design in order to find
    a solution of an inverse problem with structured output. We employ a
    Bayesian approach to handle the black-box objective function and its uncertainty.
    \item To handle the structured output, we design an acquisition function that considers the correlation between the components of the multidimensional output, and we proposed an input search algorithm based on Bayesian optimization.
    \item The effectiveness of the proposed method is verified by applying it to actual silicon carbide (SiC) crystal growth~\citep{kusunoki2014top} simulation data.
\end{itemize}

\subsection*{Related Work} 

Inverse problems can be broadly divided into two types: (i) those where the true function and output are
observable but the input is not, and (ii) those where the input and output are observable but the true function
is not. As an approach to the former type of problem, several methods based on Bayesian estimation (also
called the Bayesian inverse problem) have been studied~\citep{dashti2016bayesian, ardizzone2018analyzing}.
These approaches assume a prior distribution on the unobservable input and the goal is to
estimate the posterior distribution of $y \mid x$.
For the latter type of problem, we can utilize an emulator of the black-box function, learned by the observable input-output data~\citep{conti2009gaussian, conti2010bayesian, chen2019bayesian}. Once we have learned the emulator, we can fine-tune the input to achieve
the desired output by using sensitivity analysis. 

Data assimilation~\citep{kalnay2003atmospheric, bannister2017review} is a similar problem setting. This is particularly used to obtain an initial values that better explain the simulation results of time-evolving systems. 
In the system considered in data assimilation, in addition to the function from input $\x$ to output $\y$, the functions that describes the time evolution of $\x$ are also subject to modeling.

Bayesian optimization~\citep{brochu2010tutorial, shahriari2015taking}, and in particular that which is based on
Gaussian processes~\citep{rasmussen2006gaussian},
has been widely studied as a model-based active learning method for black-box functions.
Bayesian optimization defines an appropriate acquisition function from a model according to the specific problem, and determines the input point for the next observation as the maximizer of this function.
We consider typical acquisition functions, including optimistic ones such as the Gaussian process upper confidence bound (GPUCB), ones based on improvements such as the probability of improvement (PI) and expected improvement (EI), and ones based on information gain such as entropy search~\citep{hennig2012entropy}.

A Bayesian optimization method that has been proposed for multi-output functions with correlated structures, called multi-task Bayesian optimization~\citep{swersky2013multi}, is based on multi-output Gaussian processes~\citep{bonilla2008multi}.
Multi-output Gaussian processes can appropriately model the multiple output function with a specific structure, by constructing a covariance function of the Gaussian processes from the input similarity matrix and the output similarity matrix.
This method will be explained in detail in Section~\ref{formulation}.
Note that although in \citep{uhrenholt2019efficient}, a Bayesian 
optimization algorithm for estimating target vector-valued
output has been proposed, the structure of the output was ignored in their work.  

The rest of the paper is organized as follows. In Section~\ref{formulation},
we explain the problem setup and give some definitions. Section~\ref{main}
provides an explanation of the proposed method. The overall active learning
algorithm and two acquisition functions are also outlined in this section.
In Section~\ref{experiments} we show our experimental results using both synthetic and real datasets, in order to demonstrate the effectiveness of the proposed method.
\section{Preliminaries}
\label{formulation}

\subsection{Problem Setup}

Let $\mathcal{X} \subset \Rbb^d$ and $\mathcal{Y} \subset \Rbb^M$ be  
input and output spaces, respectively. 
For a multiple output black-box objective function
$f : \mathcal{X} \rightarrow \mathcal{Y}$, 
consider the task of searching for an input $\x_0 = (x_{0, 1}, ..., x_{0, d}) 
\in \mathcal{X}$ corresponding 
to a given desired output 
$\f_0 
= 
(f_{0, 1}, ..., f_{0, M})
=
(f_1(\x_0), ..., f_M(\x_0)) 
\in \mathcal{Y}$, where $f_m$, $m = 1, ..., M$ are scalar-valued functions 
corresponding to $m$-th element of $f$. Here we want to find $\x_0$ as few 
evaluations of $f$ as possible because $f$ can be expensive to evaluate. 
In addition, $f$ is generally observed with noise as 
\begin{align*}
    \y= (y_1, ..., y_M) = f(\x) + \boldsymbol{\varepsilon},
\end{align*}
where 
$\boldsymbol{\varepsilon} 
= 
(\varepsilon_1, ..., \varepsilon_M)  
\sim 
\mathcal{N}(\0, \Sigma)$ are independent Gaussian 
noise processes and $\Sigma$ is a diagonal matrix with elements $\sigma_m^2$, $m = 1, ..., M$. 
Therefore, we need to design a query strategy that will control the trade-off 
between exploration (sampling to reduce the uncertainty of $f$) and 
exploitation (sampling to find $\x_0$ in the region with low uncertainty). 
Since we are considering a function $f$ with a structured output, it would be natural 
to assume that $f_1, ..., f_M$ are correlated. 
To handle such a situation, $f$ is modeled by a multi-output Gaussian processes.


\subsection{Gaussian Process Modeling for Structured-Output Systems}
\label{GPModel}
To deal with the vector-valued function $f$, 
we employ the multi-output Gaussian 
process surrogate model~\citep{alvarez2012kernels}. 
For a finite number of inputs $\x_1, ..., \x_N$, 
the zero-mean Gaussian process prior is given as 
\begin{equation*}
    \underbrace{
    \left[
    \begin{array}{c}
    \f_1 \\ \f_2 \\ {\vdots} \\ \f_M
    \end{array}
    \right] 
    }_{= \f}
    \sim 
    \mathcal{N}\left(
    \left[
    \begin{array}{c}
    \0 \\ \0 \\ \vdots \\ \0
    \end{array}
    \right],
    \underbrace{
    \left[\begin{array}{cccc}
    \K_{\f_1, \f_1} & \cdots & \K_{\f_1, \f_M} \\ 
    \K_{\f_2, \f_1} & \cdots & \K_{\f_2, \f_M} \\ 
    {\vdots} & \cdots & {\vdots} \\ 
    \K_{\f_M, \f_1} & {\cdots} & \K_{\f_M, \f_M}
    \end{array}\right]
    }_{=\K}
    \right), 
\end{equation*}
where $\f_m = (f_m(\x_1), ..., f_m(\x_N))^{\top}$ is a 
vector-valued
function corresponding to $f_m$ and the entries 
$\K_{\f_m, \f_{m^{\prime}}}$ of the kernel matrix corresponding to the covariances between $\f_m$ and $\f_{m^{\prime}}$ which is induced by the 
kernel function for vector-valued output defined as 
\begin{align*}
    \K_{m, m^{\prime}}(\x, \x^{\prime}) 
    = 
    B_{m, m^{\prime}} \times k(\x, \x^{\prime}),
\end{align*} 
where $B_{m, m^{\prime}}$ describes the correlation between the function $f_m$ and 
$f_{m^{\prime}}$ ($m, m^{\prime} = 1, ..., M$), and $k$ is a kernel function on $\mathcal{X}$ such as Gaussian kernel. 
Thus, the covariance matrix $\K$ can be written as 
\begin{align*}
    \K
    &=
    \left(\begin{array}{ccc} 
    B_{1,1} \times K & {\dots} & B_{1, M} \times K \\ 
    {\vdots} & {\ddots} & {\vdots} \\ 
    B_{M, 1} \times K & {\dots} & B_{M, M} \times K
    \end{array}\right) \\ 
    &=\B \otimes K
\end{align*}
where $\B$ is $M \times M$ correlation matrix of $f_1, ..., f_M$. Hence $\K$ is $MN \times MN$ matrix. 
Then for the observations $\X = (\x_1, ..., \x_N)$, $\Y = (\y_1, ..., \y_N)$ 
and the hyperparameters of the kernel function $\thet$, 
the likelihood is given as 
\begin{align*}
    p(\Y \mid \f, \X, \thet) 
    = 
    \mathcal{N}(\Y \mid \0, \K + \boldsymbol{\Sigma}),  
\end{align*}
where $\boldsymbol{\Sigma} = \Sigma \otimes \I_N$ is an $NM \times NM$ matrix.
The predictive distribution $p(\y_* \mid \y, \X, \X_*, \thet)$
for the test input $\X_*$ is given by the Gaussian distribution 
$\mathcal{N}(\boldsymbol{\mu}_{\y_*}, \K_{\y_*})$
with the predictive mean 
\begin{align}
    \label{eq:posteriormean}
    \boldsymbol{\mu}_{\y_*} = \K_{\f_*}^{\top}(\K + \boldsymbol{\Sigma})^{-1}\y
\end{align}
and the predictive covariance matrix
\begin{align}
    \label{eq:posteriorcov}
    \K_{\y_*} = \K_{\f_*, \f_*} - \K_{\f_*}(\K + \boldsymbol{\Sigma})^{-1}\K_{\f_*}^{\top},
\end{align}
where $\K_{f_*}$ is $M \times NM$ matrix which has entries 
$K_{m, m^{\prime}}(\x_*, \x_i)$ for $i = 1, ..., N$ and 
$m, m^{\prime} = 1, ..., M$. 
\section{Bayesian Active Learning for Inverse Problems}
\label{main}
In this section, we present our proposed method, 
Bayesian active learning for inverse problems. 
In \ref{proposed}, we show the overall algorithm of the proposed method. 
Then in \ref{acquisition}, we explain the details of the 
acquisition function used in the algorithm. 
Specifically, two improvement-based acquisition functions, 
the probability of improvement~(\ref{PI}) and expected improvement~(\ref{EI}), are defined under our problem settings, 
and the probability distribution according to the objective function which is necessary to compute the acquisition functions is 
derived~(\ref{distribution}). 

\subsection{Overall Algorithm}
\label{proposed}

In the usual Bayesian optimization, the black-box function 
$f$ itself be the objective function. 
On the other hand, since our goal is to find $\x$ that achieves $\f_0$, 
the following squared error is employed as the objective function 
\begin{align}
\label{eq:objective}
    \mathcal{L}(\x) 
    = 
    \E(\x)^{\top}\E(\x)
    =
    \sum_{m = 1}^M (y_m - f_{0, m})^2, 
\end{align}
where $\E(\x) = (y_1 - f_{0, 1}, ..., y_M - f_{0, M})$. 
Then our Bayesian optimization can be formulated as an iteration of the 
following process. 
First, the next query point is determined by maximizing the acquisition 
function $\alpha$ determined from the multi-output Gaussian process surrogate model. 
Next, observe $\y$ at the specified query point. 
Finally, the observation point is added to the data set to update the 
Gaussian process. The overall procedures are shown in Algorithm~\ref{alg}. 
In the next section, we derive the two improvement-based acquisition 
functions, the probability of improvement (PI) and the expected improvement (EI), in our settings for the proposed active learning algorithm.

\begin{algorithm}[t]
\caption{Bayesian Active Learning for Inverse Problem}
\label{alg}
\begin{algorithmic}
\STATE{\bfseries Input:} observed data $\mathcal{D} = \{(\x_i, \y_i)\}_{i=1}^{N}$, 
desired output $\f_0$
\STATE{\bfseries Initialize:} $\x^* = \argmin_{\x_i, i=1, ..., N} \mathcal{L}(\x_i)$
 \FOR{$t = 1, 2, ... $} 
 \STATE{\bfseries Step~1:}  Maximize the acquisition function defined by  (\ref{eq:EI}): 
 \begin{align*}
 \x_{t+1} = \argmax_{\x \in \mX} \alpha(\x;\f_0). 
 \end{align*}
\STATE{\bfseries Step~2:} Observe $\y_{t+1} = f(\x_{t+1}) + \boldsymbol{\varepsilon}$. 
\STATE{\bfseries Step~3:} Set $\mathcal{D} \leftarrow \mathcal{D} \cup \{\x_{t+1}, \y_{t+1}\}$ and update the GP prior. 
\STATE{\bfseries Step~4:} Set $\x^* \leftarrow \argmin_{\x = \x^*, \x_{t+1}} \mathcal{L}(\x)$
 \ENDFOR
 \STATE{\bfseries Output:} $\x^*$ as an approximate of $\x_0$
\end{algorithmic}
\end{algorithm}

\subsection{The Acquisition Function}
\label{acquisition}
To minimize the squared error function $\mathcal{L}(\x)$ by 
Bayesian optimization, 
the improvement-based acquisition function such as 
probability of improvement (PI) and expected improvement (EI) 
is a natural choice. In the following, we derive those 
acquisition functions in our problem setting. 

\subsubsection{Probability of Improvement} 
\label{PI}
Let us consider the utility 
\begin{align}
    \label{eq:PIimprovement}
    u_{PI}(\x) 
    = 
    \begin{cases}
        1 & \mbox{ if } \mathcal{L}^* \ge \mathcal{L}(\x) \\
        0 & \mbox{ if } \mbox{ otherwise }
    \end{cases}
\end{align}
that returns $1$ if $\mathcal{L}(\x)$ improves from the 
current best $\mathcal{L}^* = \min_{i = 1, ..., N} 
\mathcal{L}(\x_i)$ and $0$ otherwise. 
Then the PI acquisition function is given by 
\begin{align}
    \label{eq:PI}
    \alpha_{PI}(\x) 
    &=
    \Ebb[u_{PI}(\x)] \nonumber \\ 
    &=
    {\rm Pr}(\mathcal{L}^* \ge \mathcal{L}(\x)), 
\end{align}
where the expectation $\Ebb[\cdot]$ is taken over the 
distribution of $\mathcal{L}(\x)$. 

\subsubsection{Expected Improvement} 
\label{EI}
Now consider the following utility function:
\begin{align}
    \label{eq:EIimprovement}
    u_{EI}(\x) = \max\{0, \mathcal{L}^* - \mathcal{L}(\x)\}, 
\end{align}
that returns improvement $\mathcal{L}^* - \mathcal{L}(\x)$ 
if $\mathcal{L}(\x)$ improves from the $\mathcal{L}^*$ and $0$ 
otherwise. 
The EI acquisition function in our problem is defined by 
\begin{align}
    \label{eq:EI}
    \alpha_{EI}(\x) 
    = 
    \Ebb[u_{EI}(\x)]
\end{align}
where the expectation $\Ebb[\cdot]$ is taken over the 
distribution of $\mathcal{L}(\x)$ as for PI.

In the case of ordinary Bayesian optimization where $f$ itself is 
the objective 
function, the above expectation is taken with respect to the normal 
distribution,  that is the predictive distribution of $f(\x)$. 
On the other hand, in our case, 
we need to take the expectation in (\ref{eq:PI}) 
and (\ref{eq:EI}) with respect to the distribution of 
$\mathcal{L}(\x)$ which we will discuss in the following 
section. 


\subsubsection{The Distribution of the Squared Error}
\label{distribution}
We derive the distribution of the squared error $\mathcal{L}(\x)$ 
considering the correlation of $f_1, ..., f_M$ based on the techniques from~\citep{graybill1976theory}. 

From the discussion in Section~\ref{GPModel} and \ref{proposed}, we can see 
that our target is a distribution that follows a quadratic form 
$\E(\x)^{\top}\E(\x)$ of a 
probability vector $\E(\x)$ that follows the Gaussian distribution 
defined as $\mathcal{N}(\boldsymbol{\mu}_{\y_*} - \f_0, \K_{\y_*})$, where 
$\boldsymbol{\mu}_{\y_*}$ and $\K_{\y_*}$ are the predictive mean and the 
predictive covariance matrix defined by (\ref{eq:posteriormean}) and 
(\ref{eq:posteriorcov}) respectively. 

Let 
$\d = \K_{\y_*}^{-1/2}(\E(\x) - \m)$ 
where $\m = \boldsymbol{\mu}_{\y_*} - \f_0$, then 
$\d \sim \mathcal{N} (\0, \I_M)$. 
Since $\K_{\y_*}$ is a symmetric positive semi-definite matrix, 
it can be decomposed as  
\begin{align*}
    \K_{\y_*} = \P^{\top} {\rm diag}(\lambda_1, ..., \lambda_M) \P
\end{align*}
where $\P$ is the orthogonal matrix and $\lambda_m \ge 0$, $1, ..., M$ are 
the eigenvalues of $\K_{\y_*}$. 
Let $\u = \P \d$ then $\u \sim \mathcal{N}(\0, \I_M)$ because the orthogonal 
transformation of standard normal random vector follows the same distribution. 
Then, the quadratic form $\E(\x)^{\top}\E(\x)$ can be rewritten as 
\begin{align*}
     \E(\x)^{\top}\E(\x) 
    &=
    (\d + \K_{\y_*}^{-1/2}\m)^{\top}
    \K_{\y_*}(\d + \K_{\y_*}^{-1/2}\m) \\ 
    &= 
    (\u + \b)^{\top} 
    {\rm diag} \left(\lambda_1, ..., \lambda_M \right)
    (\u + \b)
\end{align*}
where $\b = \P \K_{\y_*}^{-1/2} \m \in \Rbb^M$. 
Obviously each element $u_m + b_m$ of 
$\u + \b$ follows the normal distribution with mean 
$b_m = (\P \K_{\y_*}^{-1/2} \m)_m$ and 
variance $1$. Eventually, we can see that $\E(\x)^{\top}\E(\x)$ follows the same 
distribution that
\begin{align*}
    W = \sum_{m=1}^M \lambda_m w_m
\end{align*}
follows. Here $w_m$, $m = 1, ..., M$ 
follows the independent non-central $\chi^2$ distribution with one degree 
of freedom and non-centrality parameter $b_m^2$. 
Such a distribution is known as generalized $\chi^2$ 
distribution~\citep{imhof1961computing}. 
Hence, the PI acquisition function (\ref{eq:PI}) is the cumulative distribution 
function of the generalized $\chi^2$ distribution. 
For the EI acquisition function (\ref{eq:EI}), with the change of variables as
$\mathcal{L}(\x) = t$, we have 
\begin{align}
\label{eq:EItransform}
    \alpha(\x) 
    &= 
    \Ebb[\max\{0, \mathcal{L}^* - \mathcal{L}(\x) \}] \nonumber \\ 
    &= 
    \int_0^{\mathcal{L}^*} (\mathcal{L}^* - t) p_{G_{\chi^2}}(t) {\rm d}t \nonumber \\ 
    &=
    \mathcal{L}^* G_{\chi^2}(\mathcal{L}^*) 
        - \int_0^{\mathcal{L}^*} t p_{G_{\chi^2}}(t) {\rm d}t
\end{align}
where $p_{G_{\chi^2}}$ and $G_{\chi^2}$ are probability density function and 
cumulative distribution function of the generalized $\chi^2$ distribution, 
respectively. Although the second term in (\ref{eq:EItransform}) does not have 
a closed form, it can be transformed using partial integration techniques as
\begin{align*}
    \int_0^{\mathcal{L}^*} t p_{G_{\chi^2}}(t) {\rm d}t
    &=
    [t G_{\chi^2}(t)]_0^{\mathcal{L}^*} 
        - \int_0^{\mathcal{L}^*} G_{\chi^2}(t) {\rm d}t \\ 
    &=
    \mathcal{L}^* G_{\chi^2}(\mathcal{L}^*)
        - \int_0^{\mathcal{L}^*} G_{\chi^2}(t) {\rm d}t.
\end{align*}
Combined with (\ref{eq:EItransform}), we obtain the expression of EI as 
\begin{align}
    \label{eq:EIfinal}
    \alpha(\x) = \int_0^{\mathcal{L}^*} G_{\chi^2}(t) {\rm d}t. 
\end{align}
Since $G_{\chi^2}(t)$ is continuous and monotonically increasing function, the 
above definite integral can easily be computed by quadrature by parts on the 
closed interval $[0, \mathcal{L}^*]$. 
In the implementation, several algorithms for calculating the generalized chi-square distribution have been proposed~\citep{imhof1961computing, davies1980distribution}. 
In our experiments, we employ a computation method based on 
Edgeworth expansion~\citep{bickel1974edgeworth}. 
This method is implemented, for example, in the sadists package~\citep{sadists-Manual}.

\section{Numerical Experiments}
\label{experiments}
In this section, the effectiveness of the proposed method is demonstrated by three numerical experiments: two synthetic data experiments and one real data experiment. 
The synthetic data experiments deal with the problem of finding the 
desired ``shape" such as triangle and sphere. 
In the real data experiment, we consider the crystal growth modeling of silicon carbide~\citep{kusunoki2014top} which is a problem of materials informatics.

\subsection{Demonstration with Synthetic Data}
\begin{figure}[t]
 \begin{center}
 \begin{tabular}{cc}
 \hspace{-1.0cm}
  \includegraphics[bb=0 0 710 390, scale=0.34]{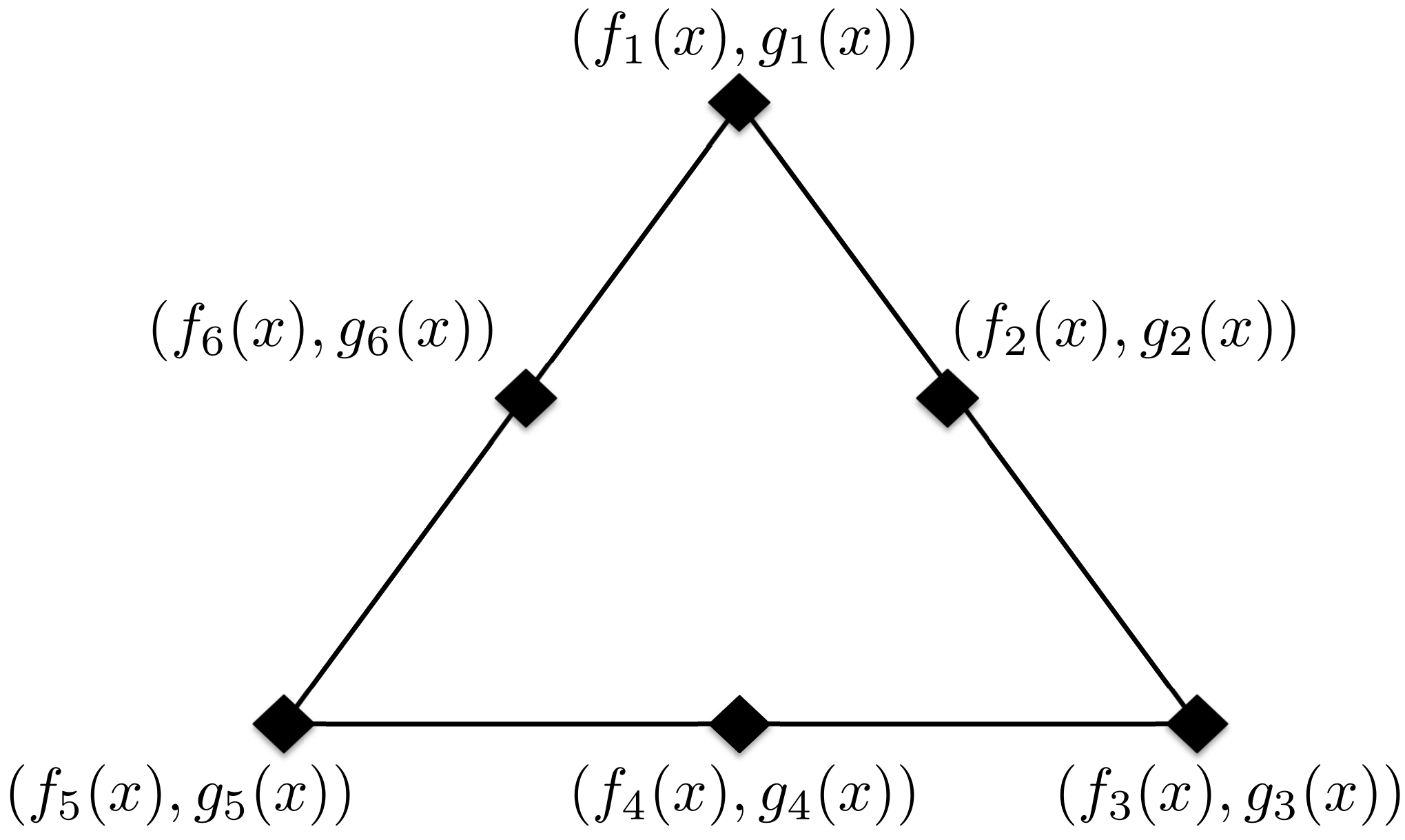} &
  \hspace{-1.5cm}
  \includegraphics[bb=0 0 586 350, scale=0.29]{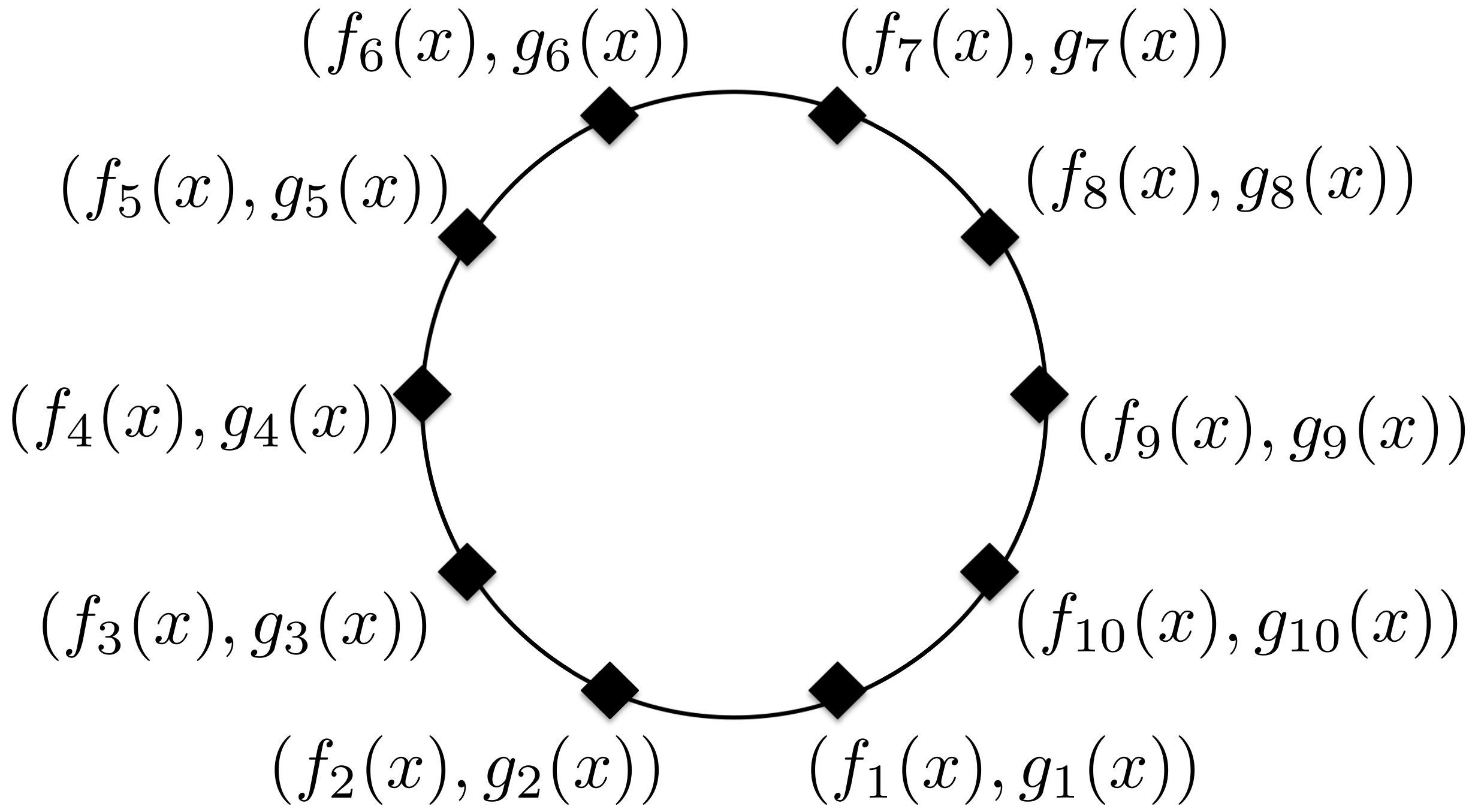} 
 \end{tabular}
 \caption{
 {\footnotesize
 Oracles in the synthetic data experiments. 
 Each has two objective functions for each two-dimensional coordinate. 
 Top: Triangle shape finding problem. Overall, it is a multiple output function with 12 objective functions. Bottom: Sphere shape finding problem. Overall, it is a multiple output function with 20 objective functions.
 }
 \label{fig:synthetic_oracle}
 }
 \end{center}
\end{figure}

\begin{figure}[t]
 \begin{center}
 \begin{tabular}{cc}
 \hspace{-1.0cm}
  \includegraphics[bb = 0 0 431 287, scale=0.55]{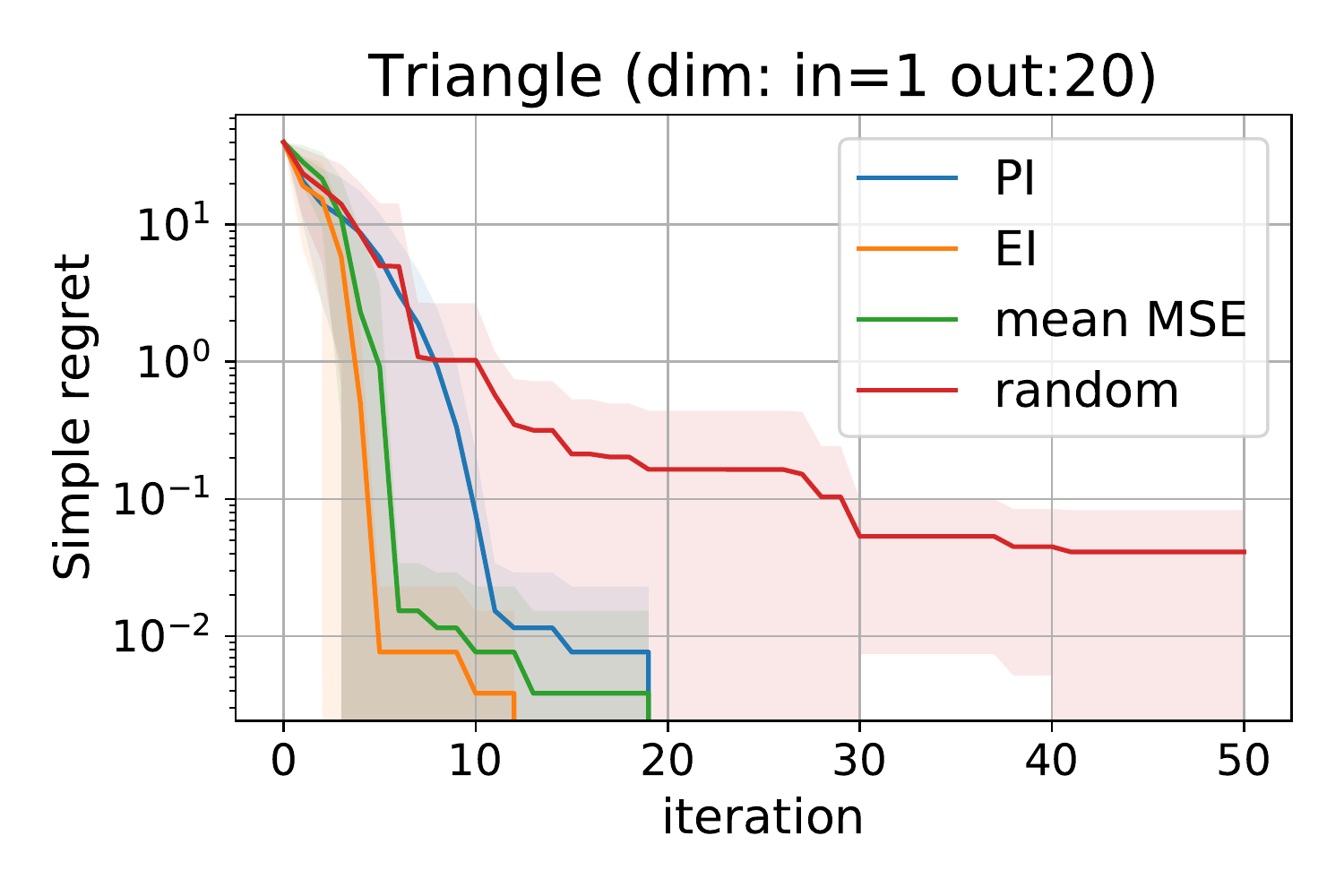} &
  \hspace{-0.7cm}
  \includegraphics[bb = 0 0 431 287, scale=0.55]{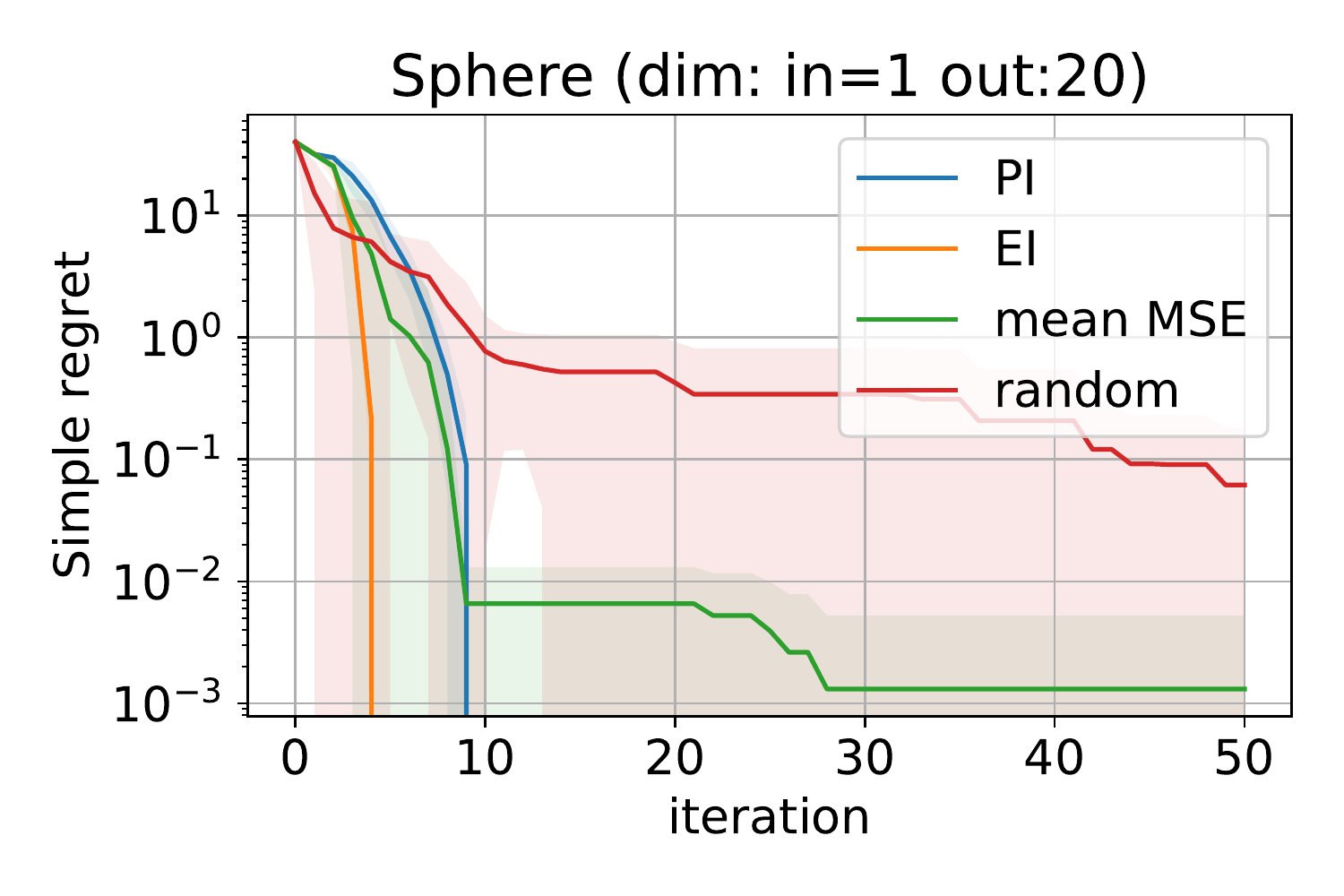} 
 \end{tabular}
 \caption{
 {\footnotesize
 Comparison results of log-scale simple regret of four methods. 
 The solid line and the shaded area represent the mean and standard deviation of 10 trials, respectively. Top: the result 
 of the Triangle-shape finding problem. Bottom: the result of the Sphere-shape finding problem. Both experiments show that the proposed method can achieve the desired output most efficiently.
 }
 \label{fig:regret_artificial}
 }
 \end{center}
\end{figure}

This section shows two demonstrations of the proposed method 
using synthetic shape finding problems. 

\subsection*{Triangle shape finding problem} 

We firstly consider the problem of finding 
an input that achieves the desired triangle in two dimensional space. 
The synthetic data are created as follows: 
We prepared oracle functions for each $x$-axis and $y$-axis 
of $6$ points corresponding to each vertex and midpoint between vertexes of 
the triangle (see top of Figure~\ref{fig:synthetic_oracle}).
Then, 100 points as pooled dataset were sampled from the oracle functions, and 2 points from 
them were used as initial dataset. Here, each data point corresponds to a $12$-dimensional 
vector of $f_i(x)$ and $g_i(x)$, $i = 1, 2, ..., 6$. The details are 
described in Appendix~\ref{oracle_detail}. 
In the experiment, we compared four methods: random sampling (random), search using only mean of the Gaussian process (mean MSE) and the proposed methods (PI and EI).

The results are shown in Figure~\ref{fig:demo_intro} and the top of 
Figure~\ref{fig:regret_artificial}. Figure~\ref{fig:demo_intro} shows a 
process from step 1 (Top left) to step 8 (Bottom right). We can see how the 
model prediction (blue triangle with error bar) 
and the actual observation (green triangle) gradually approach 
the desired output (red triangle). Finally the algorithm found the desired output in seventh round. 
Figure~\ref{fig:regret_artificial} shows the comparison of the simple regret 
among the four methods. The results show the log-scale mean and standard 
deviation over 10 trials for each method. Obviously the proposed EI outperforms the other methods and the proposed PI is comparable to the mean MSE. Further results are shown in Appendix~\ref{triangle_results}. 

\subsection*{Shpere shape finding problem} 

Second, we consider the problem of finding 
an input that achieves the desired sphere in two dimensional space. 
As in the case of the triangle, we prepared oracle functions corresponding to 
$x$ and $y$-axis of 10 points at equal intervals on the sphere (see bottom of Figure~\ref{fig:synthetic_oracle}). 
Hence we observe the $20$-dimensional vector of $f_i(x)$ and $g_i(x)$, $i = 1, 2, ..., 10$ as data. The details are 
described in Appendix~\ref{oracle_detail}. 
The other settings are the same as for the triangle.

The results are shown in the bottom of Figure~\ref{fig:regret_artificial} and
Figure~\ref{fig:results_circle}. 
As is the case with triangle, Figure 3 shows that the proposed method requires smaller 
number of iteration to reduce regrets compared to other methods. 
Figure~\ref{fig:results_circle} shows a search
process from step 1 (Top left) to step 5 (Bottom right). 
The first and third rows are the similar plots as the Figure~\ref{fig:demo_intro}. 
In the second and fourth rows, the top plot shows the model behavior and the bottom plot shows the EI acquisition function. In this case, we can see that the desired output is found in the fifth observation.

\subsection{Real Data Analysis}
\begin{figure}[t]
 \begin{center}
 \begin{tabular}{c}
 \hspace{-0.5cm}
  \includegraphics[bb=0 0 592 205, scale=0.35]{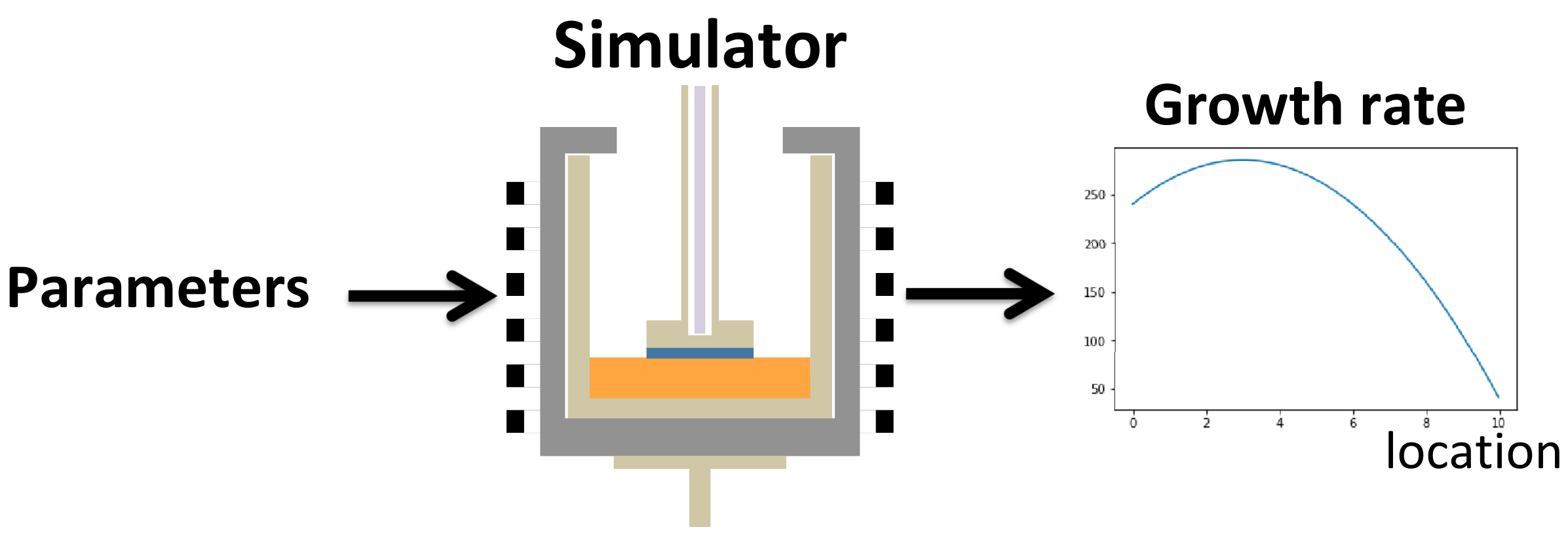}
 \end{tabular}
 \caption{
 \label{fig:SiCsim} 
 {\footnotesize
 SiC crystal growth modeling data. Since the simulator is a black box, we can only observe the pairs of input parameters and structured-outputs.
 }
 }
 \end{center}
\end{figure}

In this section, we apply the proposed method to the crystal growth modeling data 
of Silicon Carbide (SiC)~\citep{kusunoki2014top, tsunooka2018high}. This dataset consists of the input-output pairs. The input is parameter candidates that will be input to the SiC crystal growth model simulator, and the output is the crystal growth rate that is returned  from the simulator. 
We have nine kinds of input parameter candidates that control the behavior of experimental equipment: 
thermal conductivity of felt, graphite and insulator, 
electrical conductivity of graphite and insulator, 
emissivity of solution, 
heat capacity of solution, 
reaction rate constant at crystal-solution interface, 
reaction rate constant at graphite-solution interface. 
In this experiment, we focus on the reaction rate constant at graphite-solution interface and consider this one-dimensional input. 
This is because empirically, the output largely vary due to the fluctuation of reaction rate constant at graphite-solution interface, that it is difficult to determine this parameter compared to other parameters.
In addition, since the output, crystal growth rate, is measured at 20 locations, it is a multidimensional output observed as a 20-dimensional vector. 
Furthermore, since the growth rates at a near locations are expected to take a close value, this data can be regarded as a structured-output in which each dimension of the output vector has a correlation. 
Figure~\ref{fig:SiCsim} shows the data generation mechanism. 
Our goal is to quickly find the parameters of the simulator that outputs the desired growth rate vector.

\subsection*{Results} 

\begin{figure}[t]
 \begin{center}
 \begin{tabular}{c}
 \vspace{-0.3cm}
 \hspace{-0.5cm}
  \includegraphics[bb=0 0 431 287, scale=0.55]{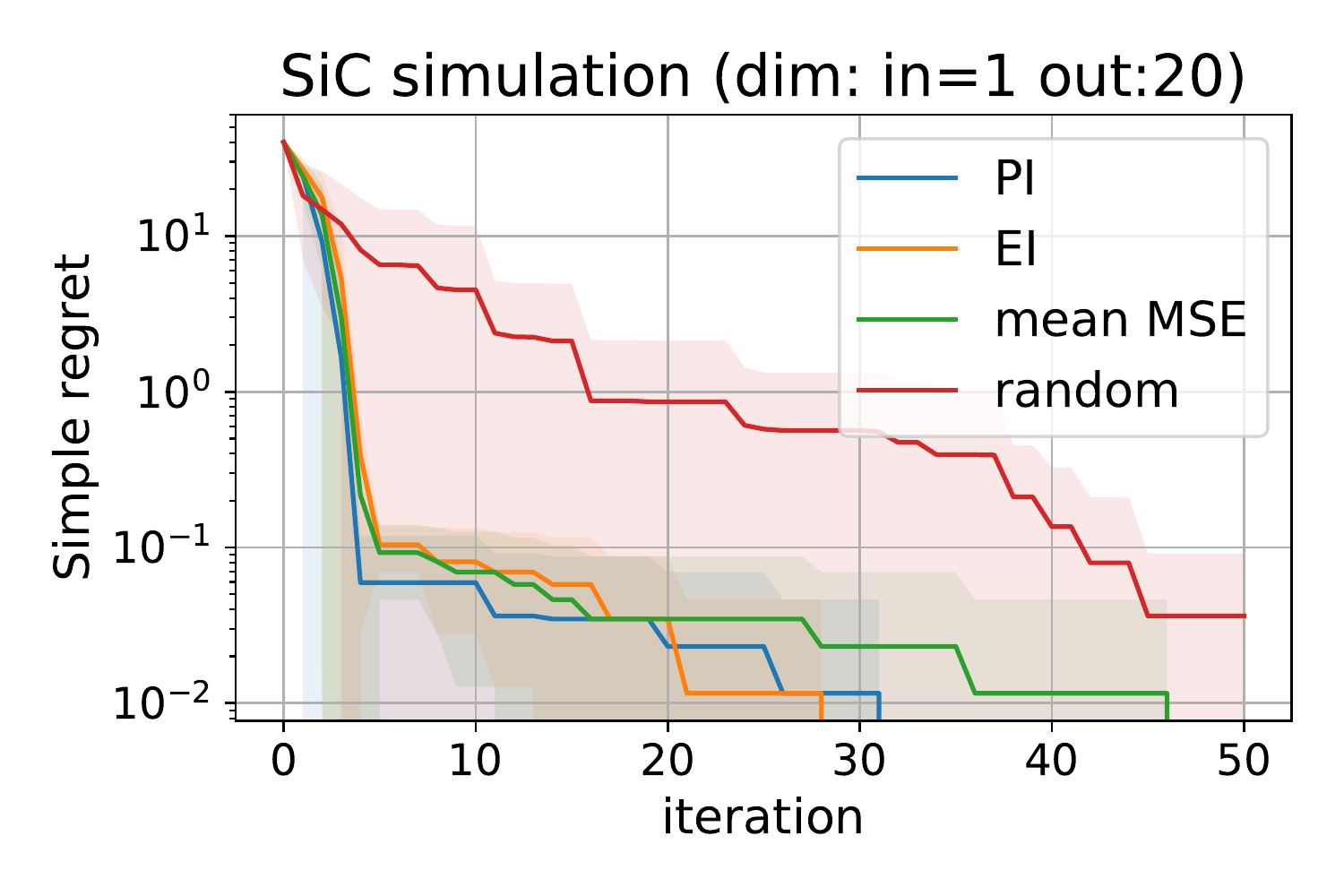}
 \end{tabular}
 \caption{
 \label{fig:SiC_regret} 
 {\footnotesize
 Comparison results of log-scale simple regret of four methods. The solid line and the shaded area represent the mean and standard deviation of 10 trials, respectively. We can see that the proposed method can achieve the desired output most efficiently.
 }
 }
 \end{center}
\end{figure}

\begin{figure}[t]
 \begin{center}
 \begin{tabular}{cccc}
 \hspace{-1.5cm}
  \includegraphics[bb=0 0 460 345, scale=0.27]{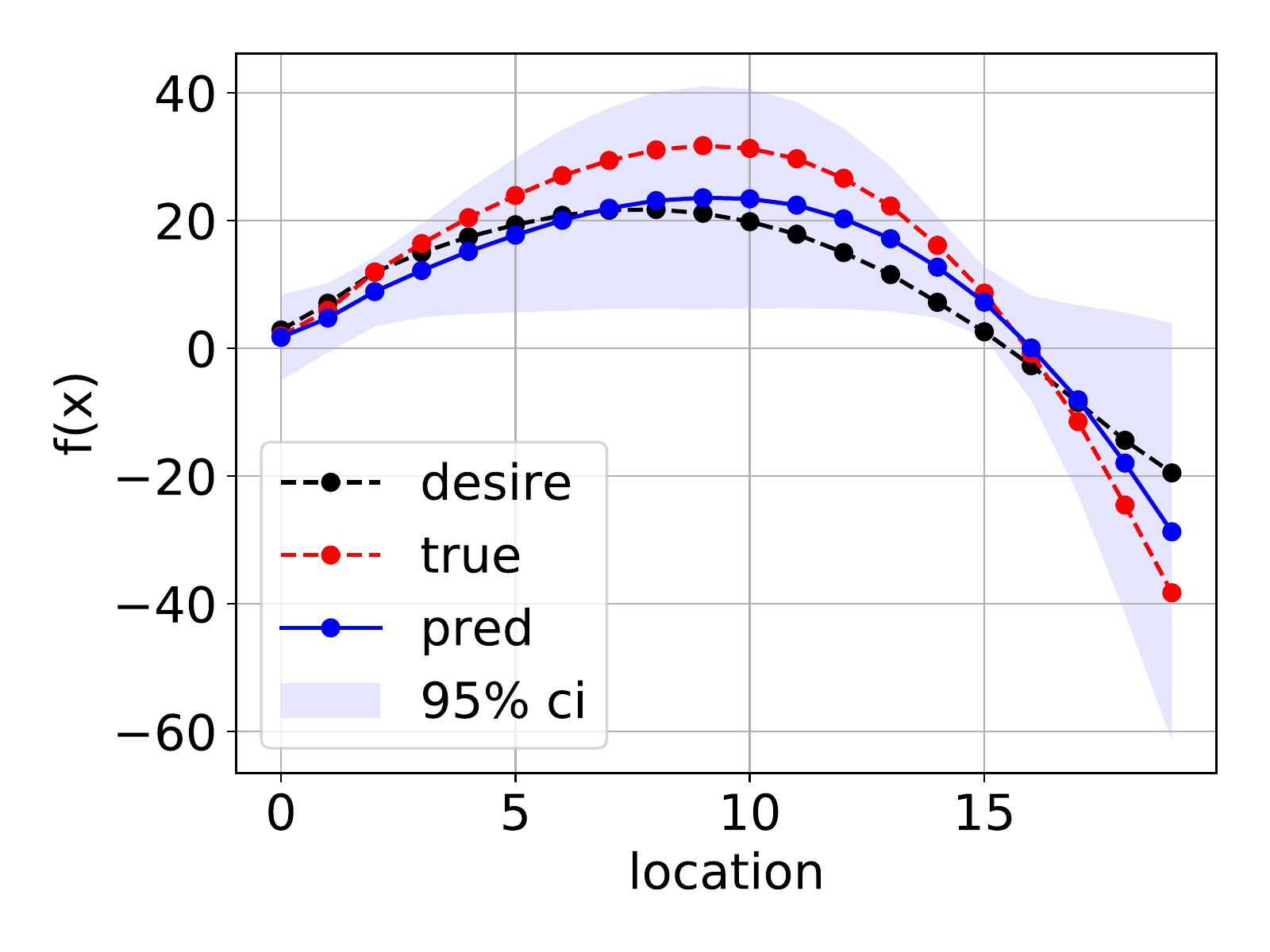} & 
  \hspace{-0.6cm}
  \includegraphics[bb=0 0 460 345, scale=0.27]{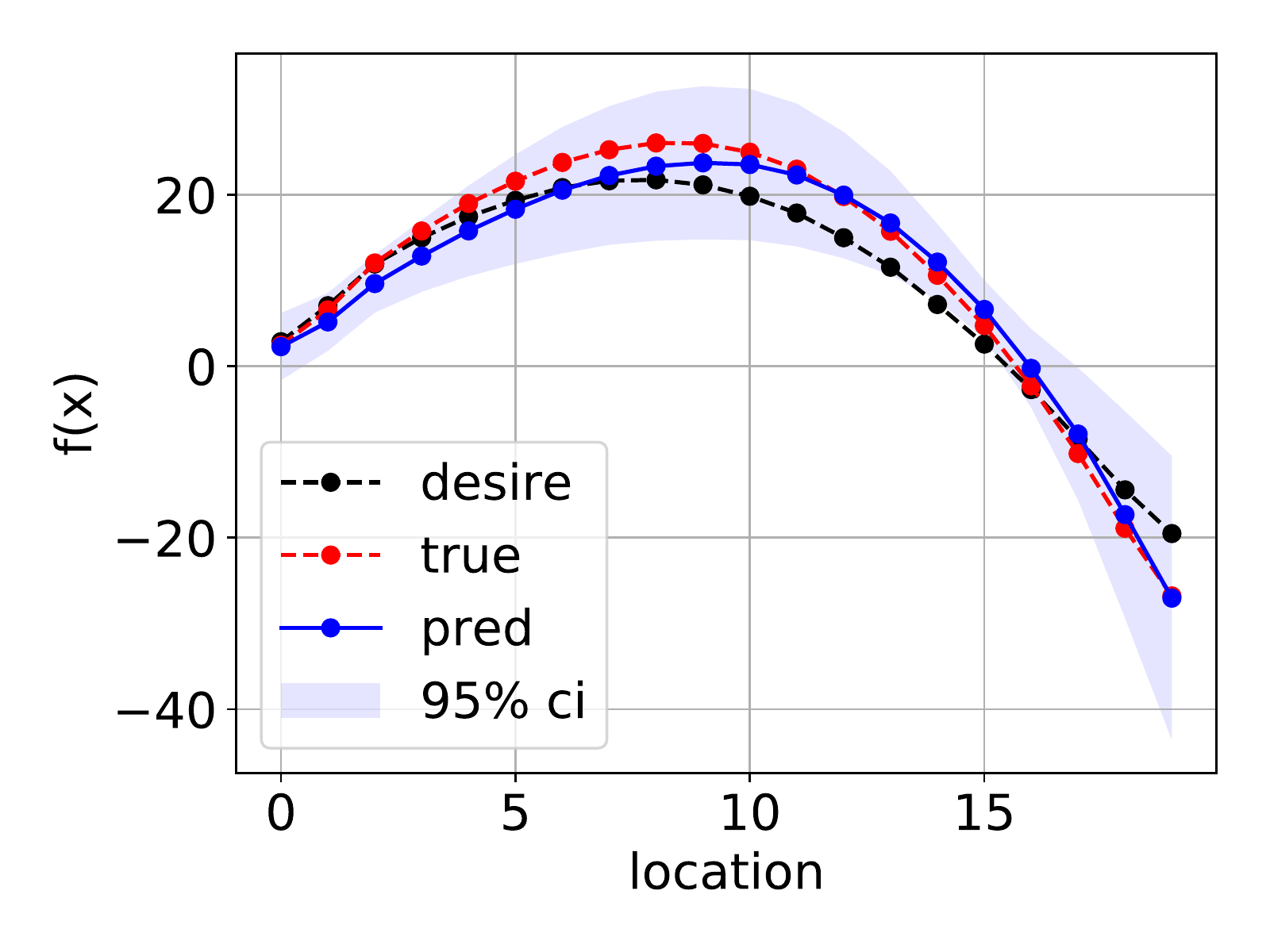} & 
  \hspace{-0.6cm}
  \includegraphics[bb=0 0 460 345, scale=0.27]{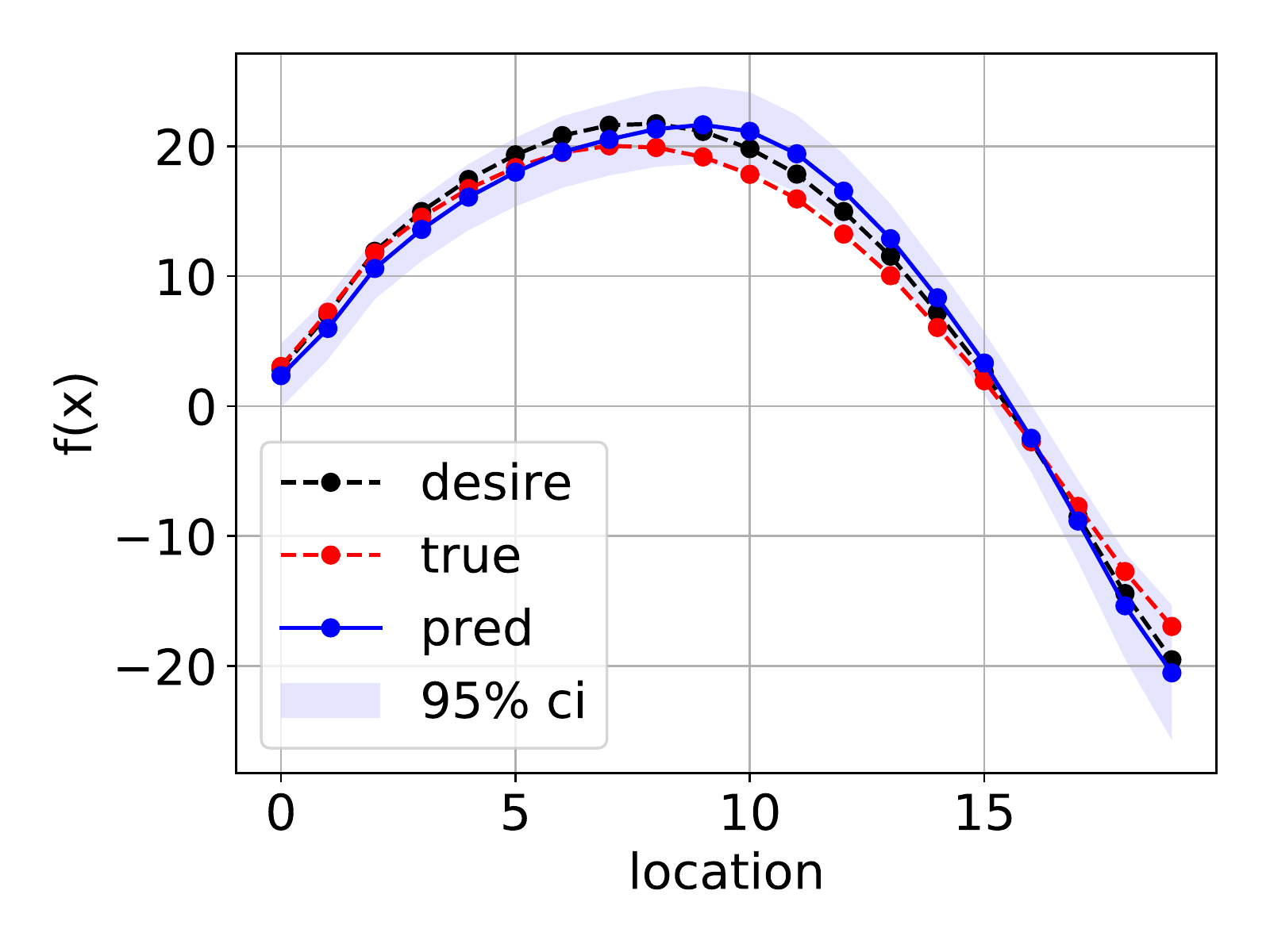} & 
  \hspace{-0.6cm}
  \includegraphics[bb=0 0 460 345, scale=0.27]{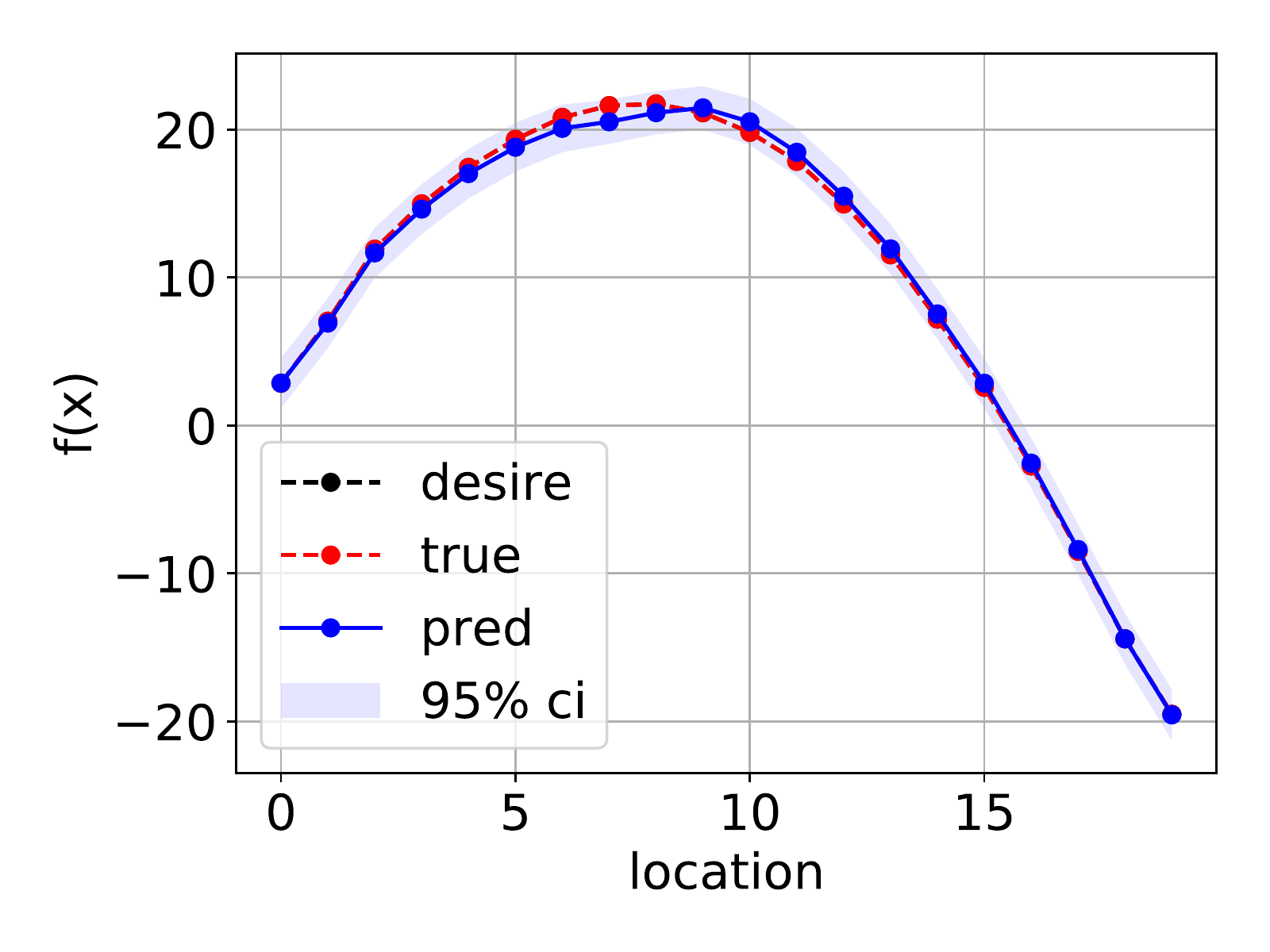} \\ 
  \hspace{-1.5cm}
  \includegraphics[bb=0 0 460 345, scale=0.27]{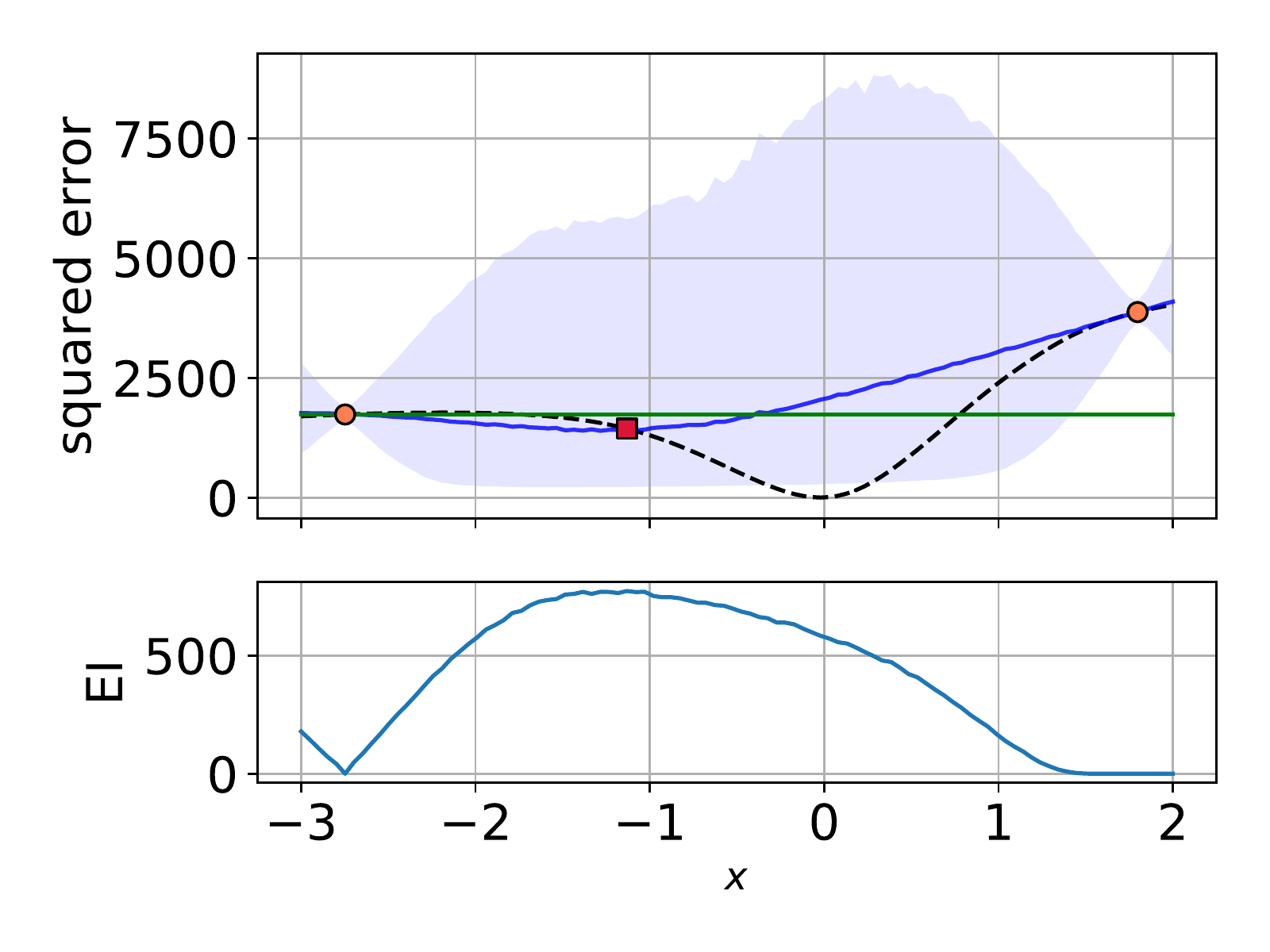} & 
  \hspace{-0.6cm}
  \includegraphics[bb=0 0 460 345, scale=0.27]{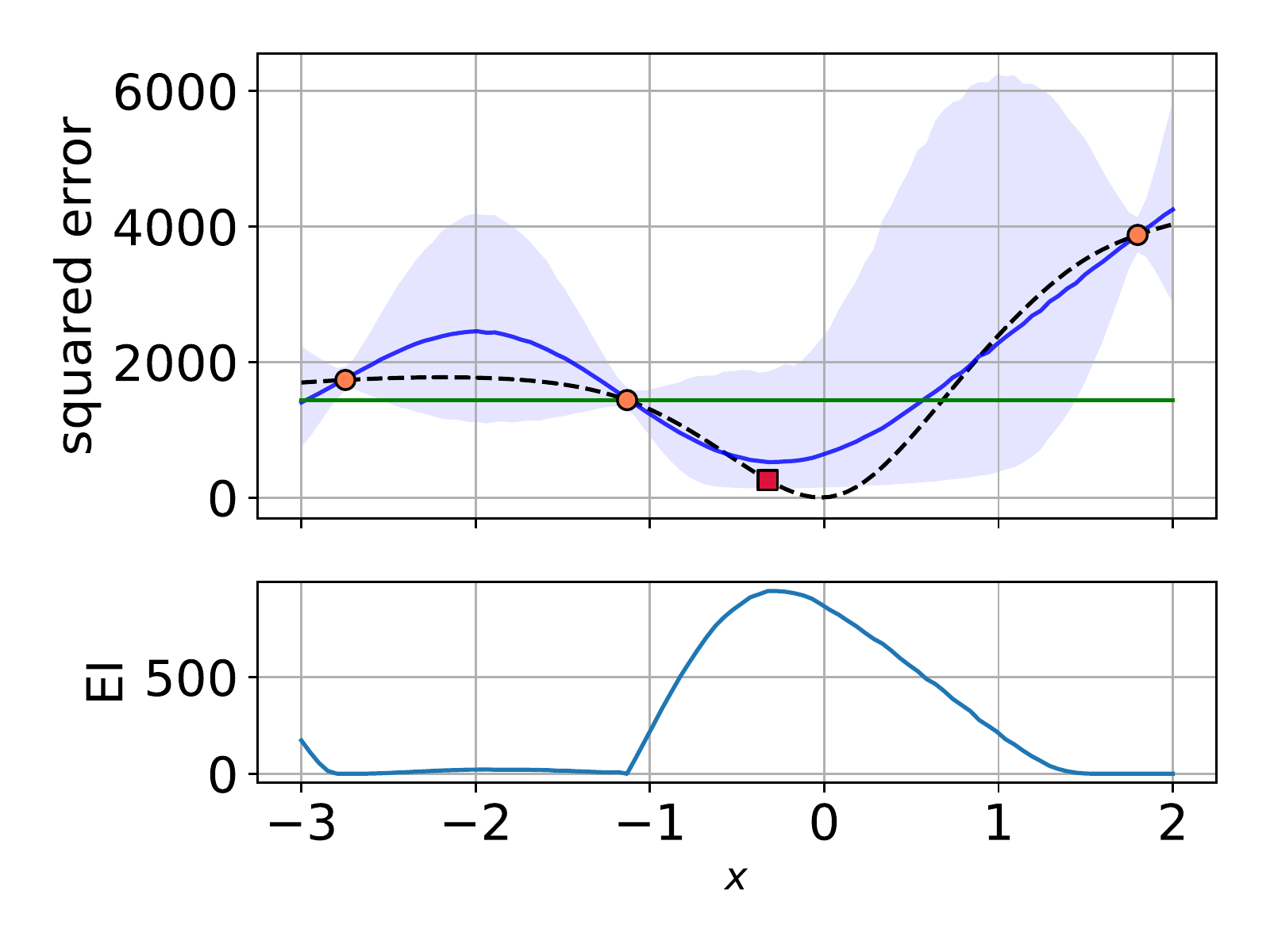} & 
  \hspace{-0.6cm}
  \includegraphics[bb=0 0 460 345, scale=0.27]{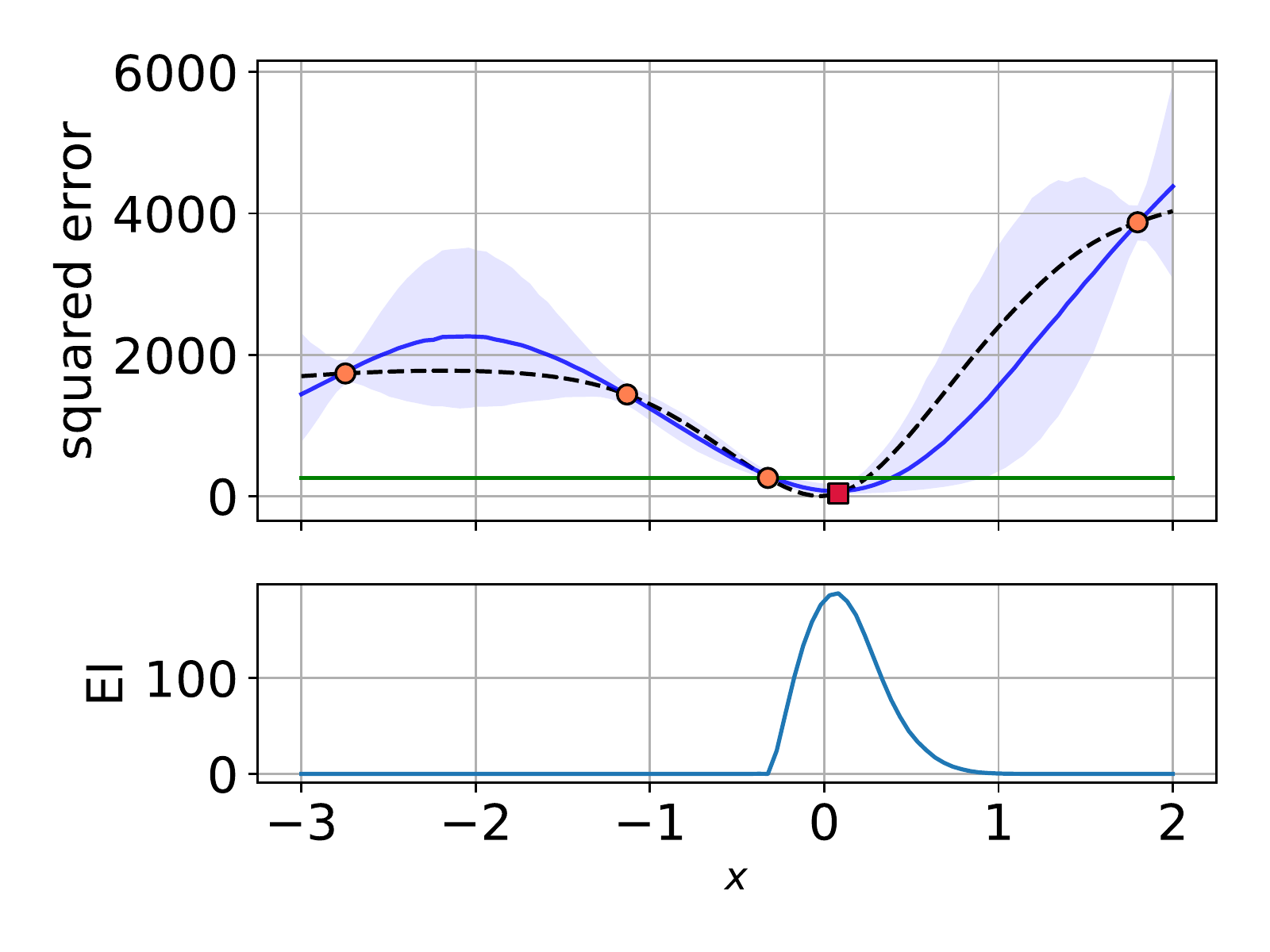} & 
  \hspace{-0.6cm}
  \includegraphics[bb=0 0 460 345, scale=0.27]{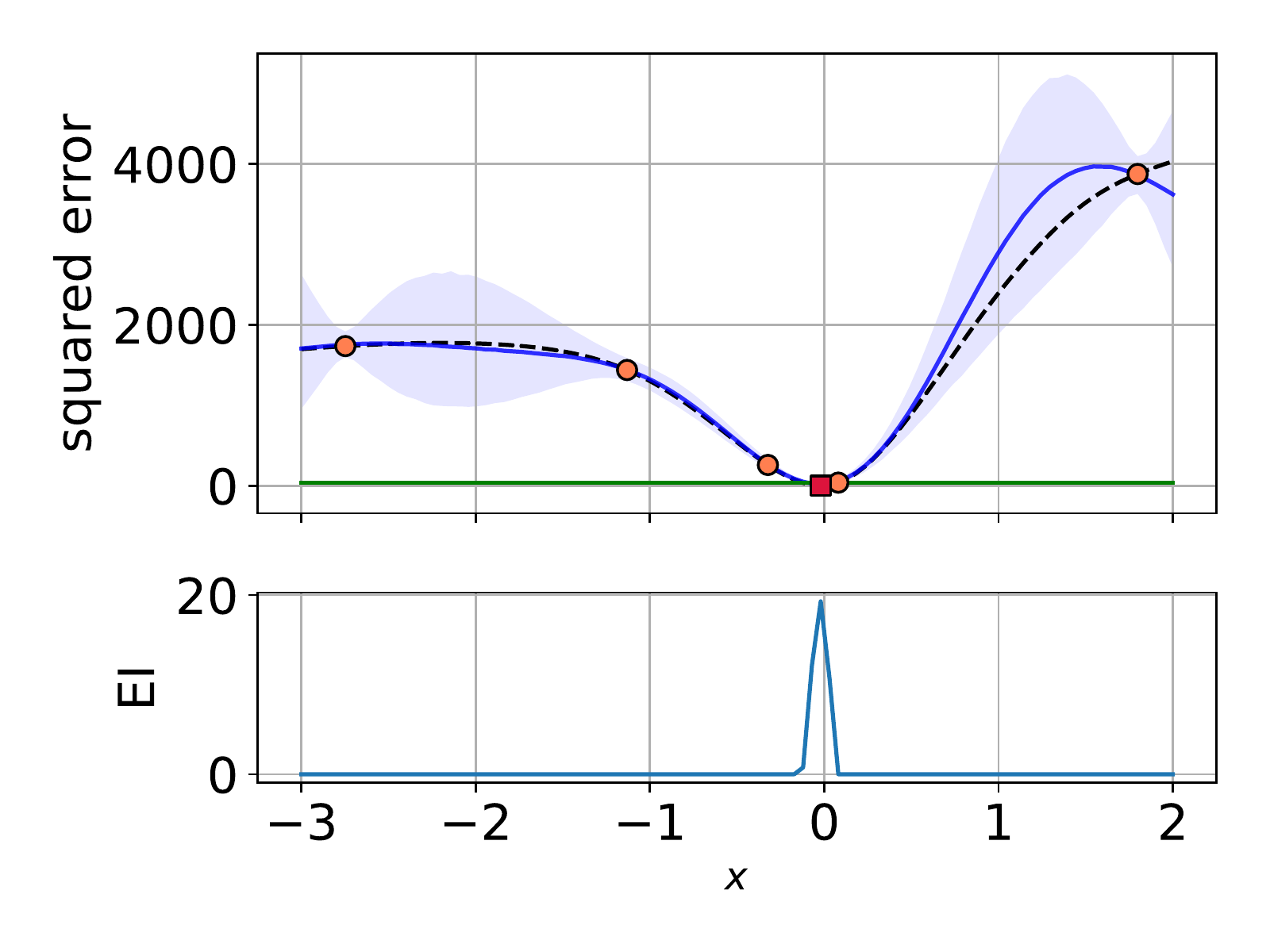}
 \end{tabular}
 \caption{
 \label{fig:SiC_plot} 
 {\footnotesize
 Shape-finding procedure and the model behavior of the proposed method using the EI acquisition function for the 
 SiC data. Top row shows the search process in the shape space and bottom row 
 shows the model fitting processes. 
 }
 }
 \end{center}
\end{figure}

The results are shown in Figures~\ref{fig:SiC_regret} and 
\ref{fig:SiC_plot}. Figure~\ref{fig:SiC_regret} shows 
a comparison of log-scale simple regret of the four methods. 
We can see that the proposed methods reduce the simple regret with fewer observations than other methods.
Figure~\ref{fig:SiC_plot} shows a search process. 
The search proceeds from the top to the bottom. 
In this case, the desired growth rate vector can be found by the fourth observation.
These experimental results imply that the proposed method can work effectively in the real structured-output search problems.

\begin{figure*}[ht]
 \begin{center}
 \begin{tabular}{ccc}
 \hspace{-0.6cm}
  \includegraphics[bb=0 0 460 345, scale=0.32]{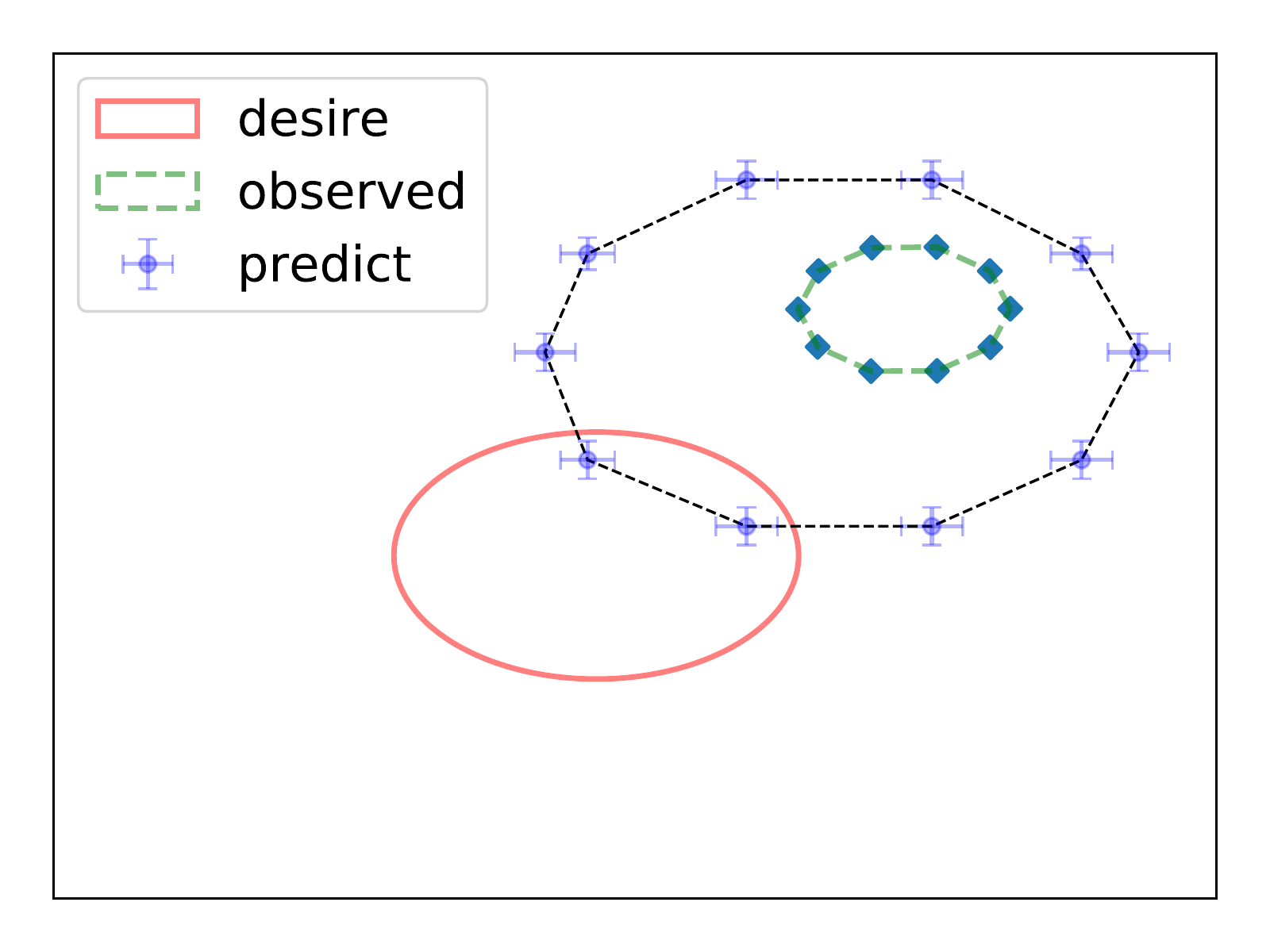} &
  \hspace{-0.6cm}
  \includegraphics[bb=0 0 460 345, scale=0.32]{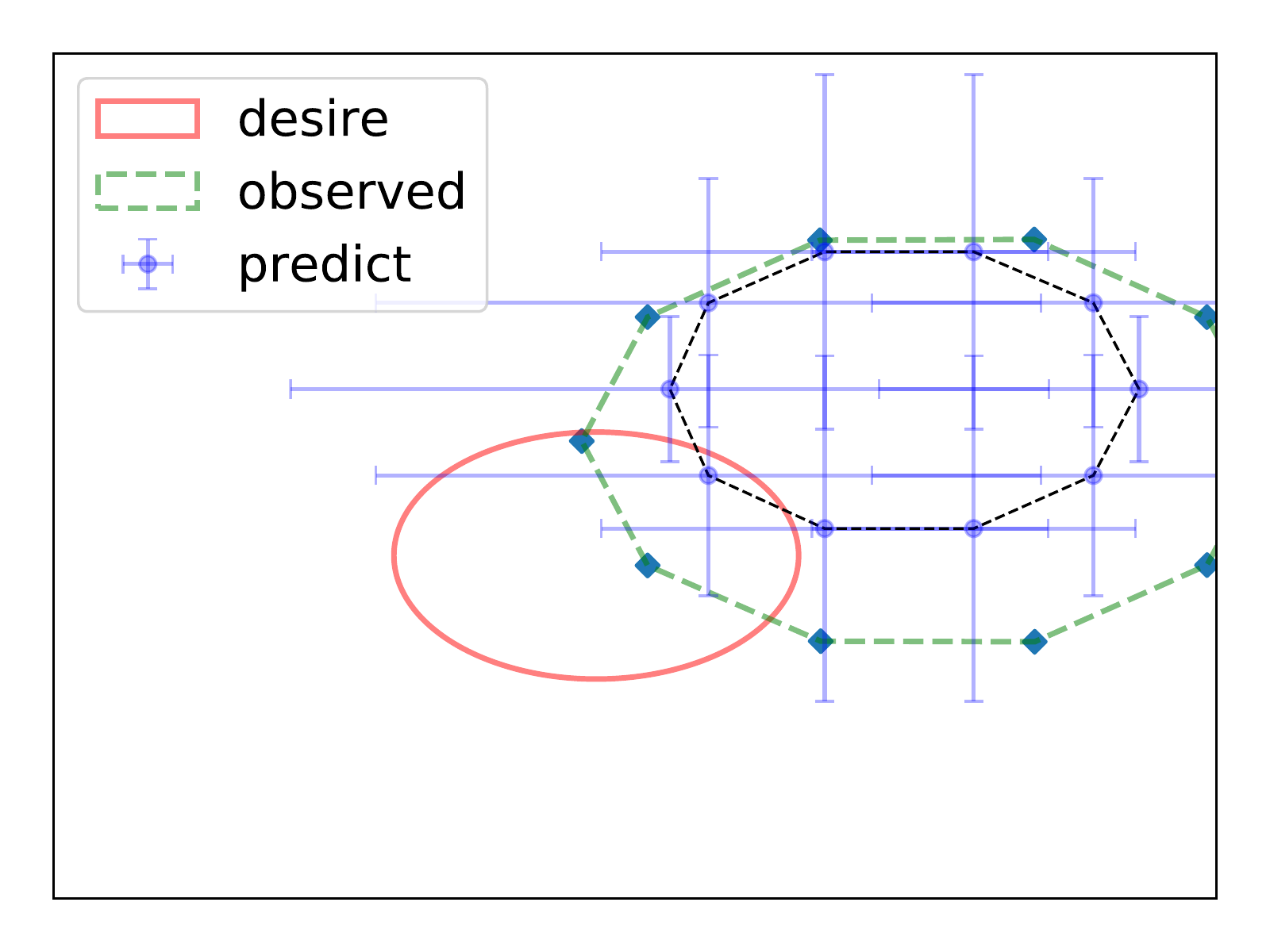}& 
  \hspace{-0.6cm}
  \includegraphics[bb=0 0 460 345, scale=0.32]{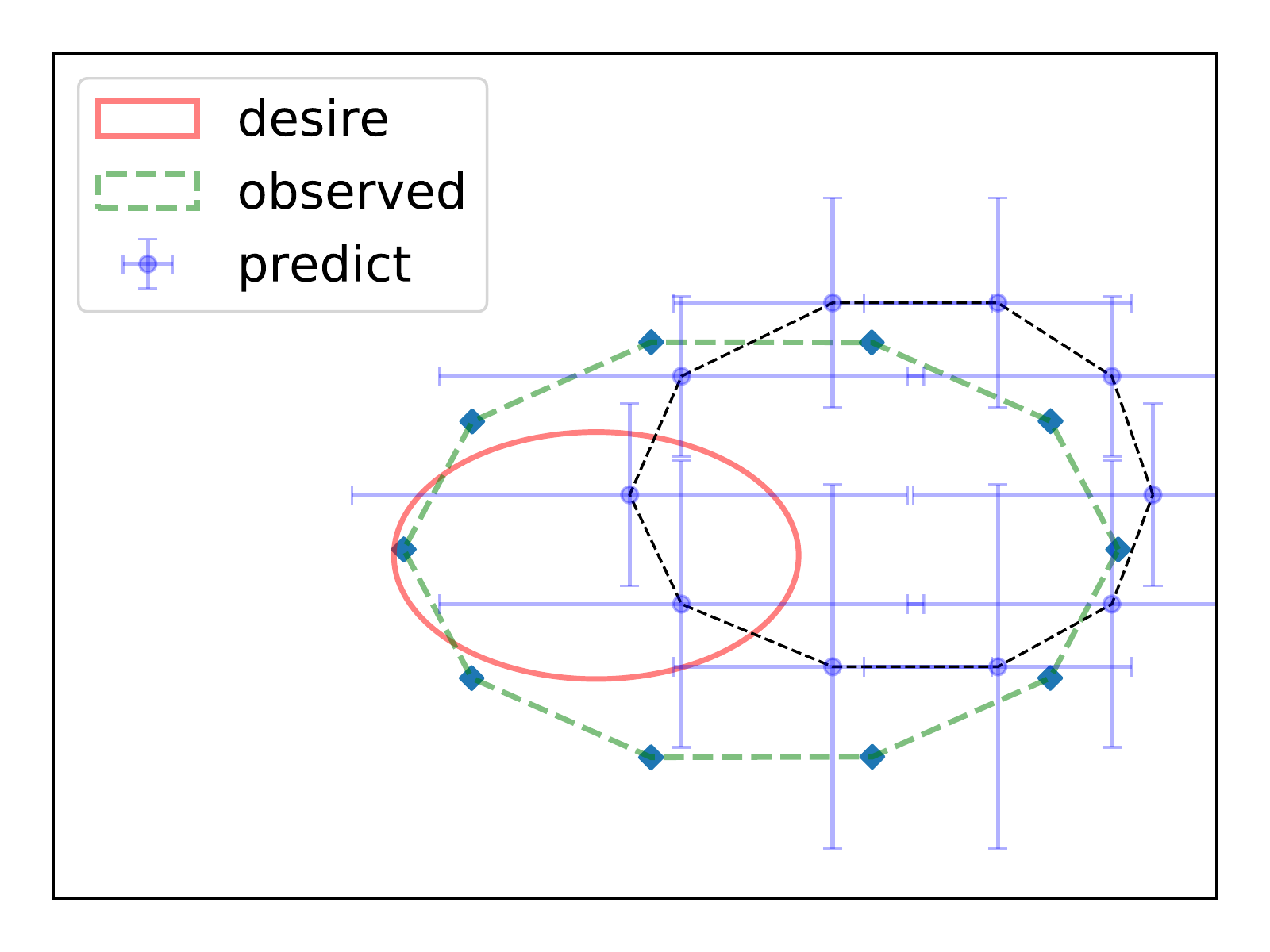} \\
  \hspace{-0.6cm}
  \includegraphics[bb=0 0 460 345, scale=0.32]{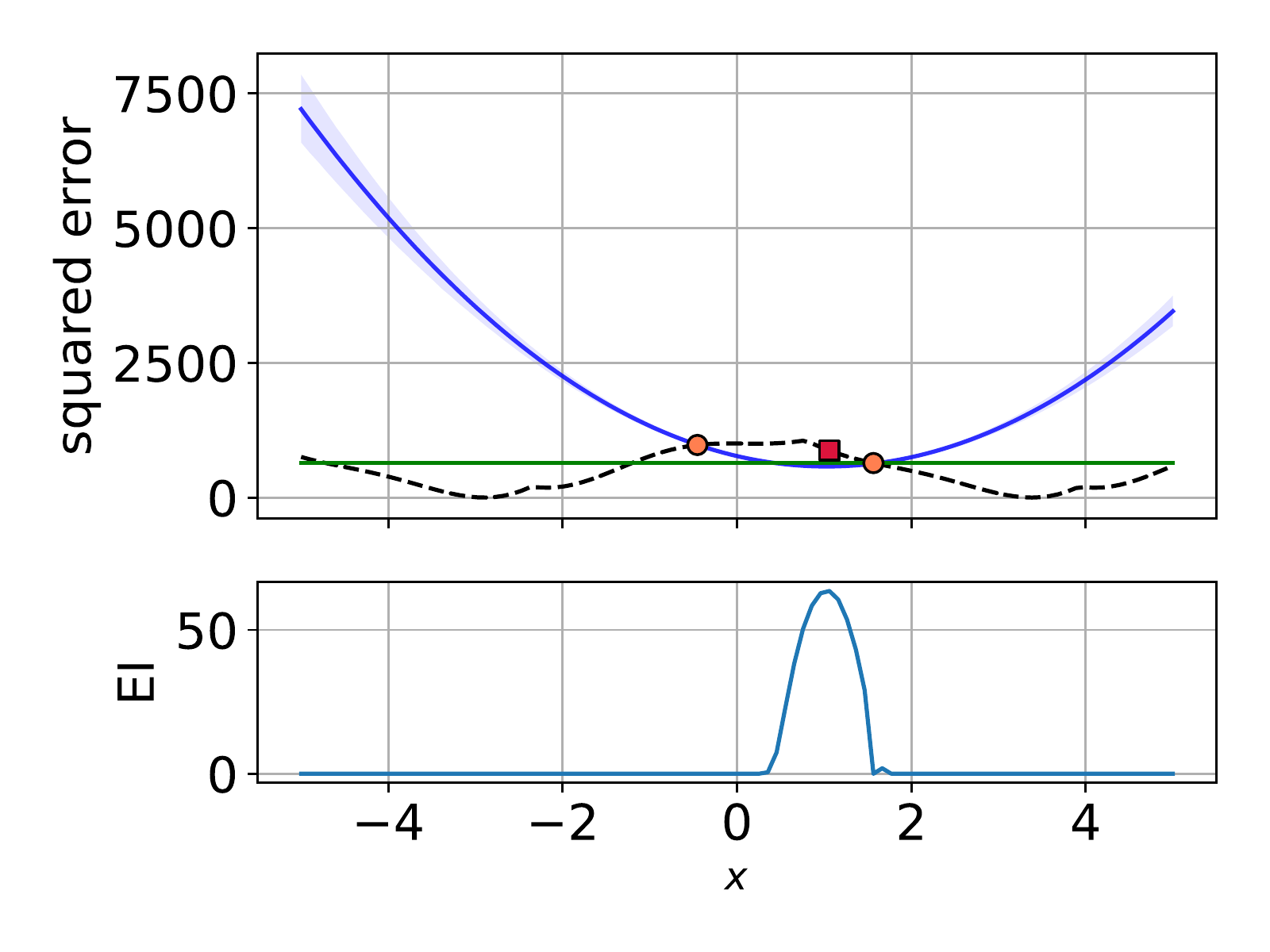} & 
  \hspace{-0.6cm}
  \includegraphics[bb=0 0 460 345, scale=0.32]{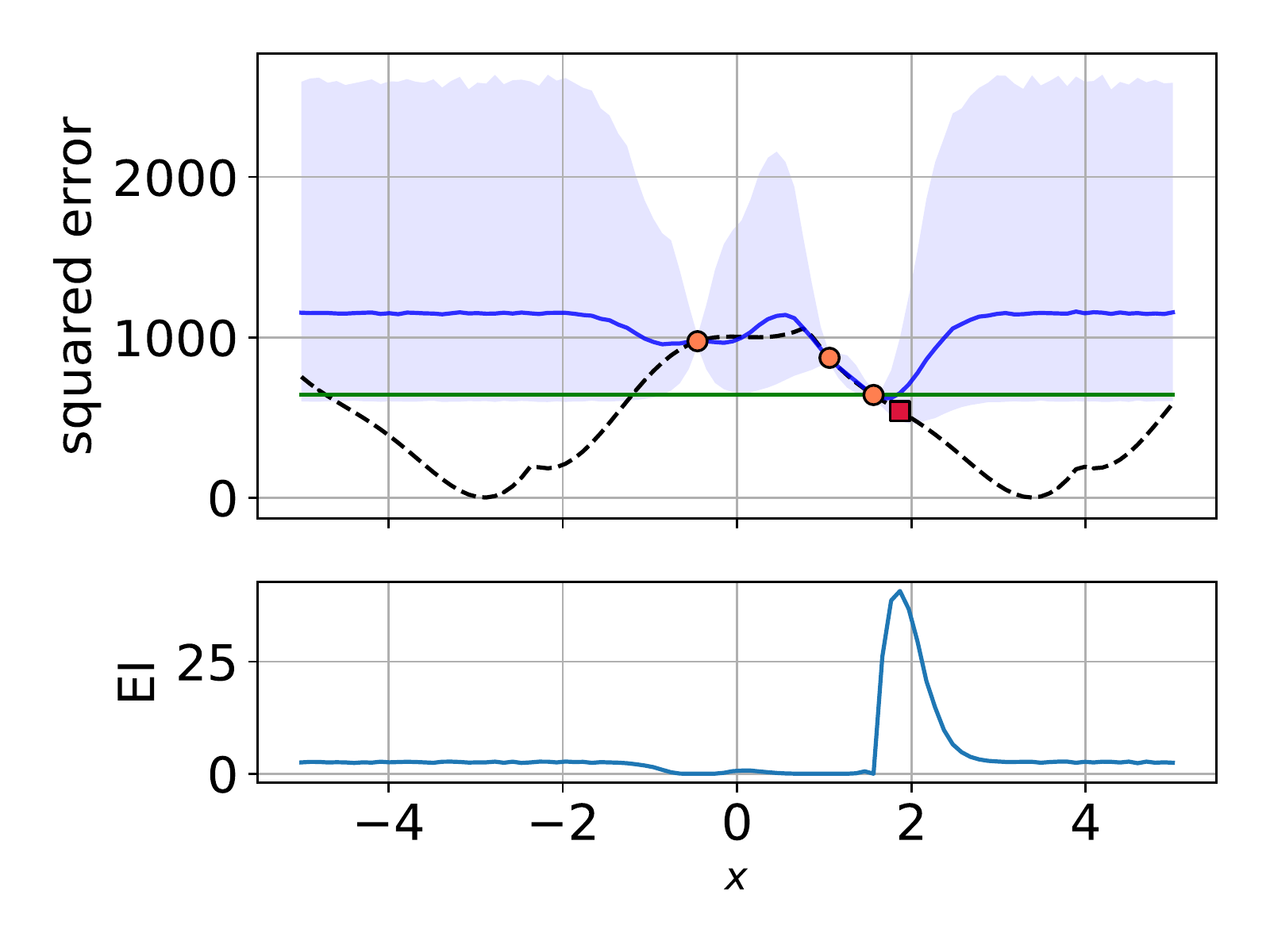} & 
  \hspace{-0.6cm}
  \includegraphics[bb=0 0 460 345, scale=0.32]{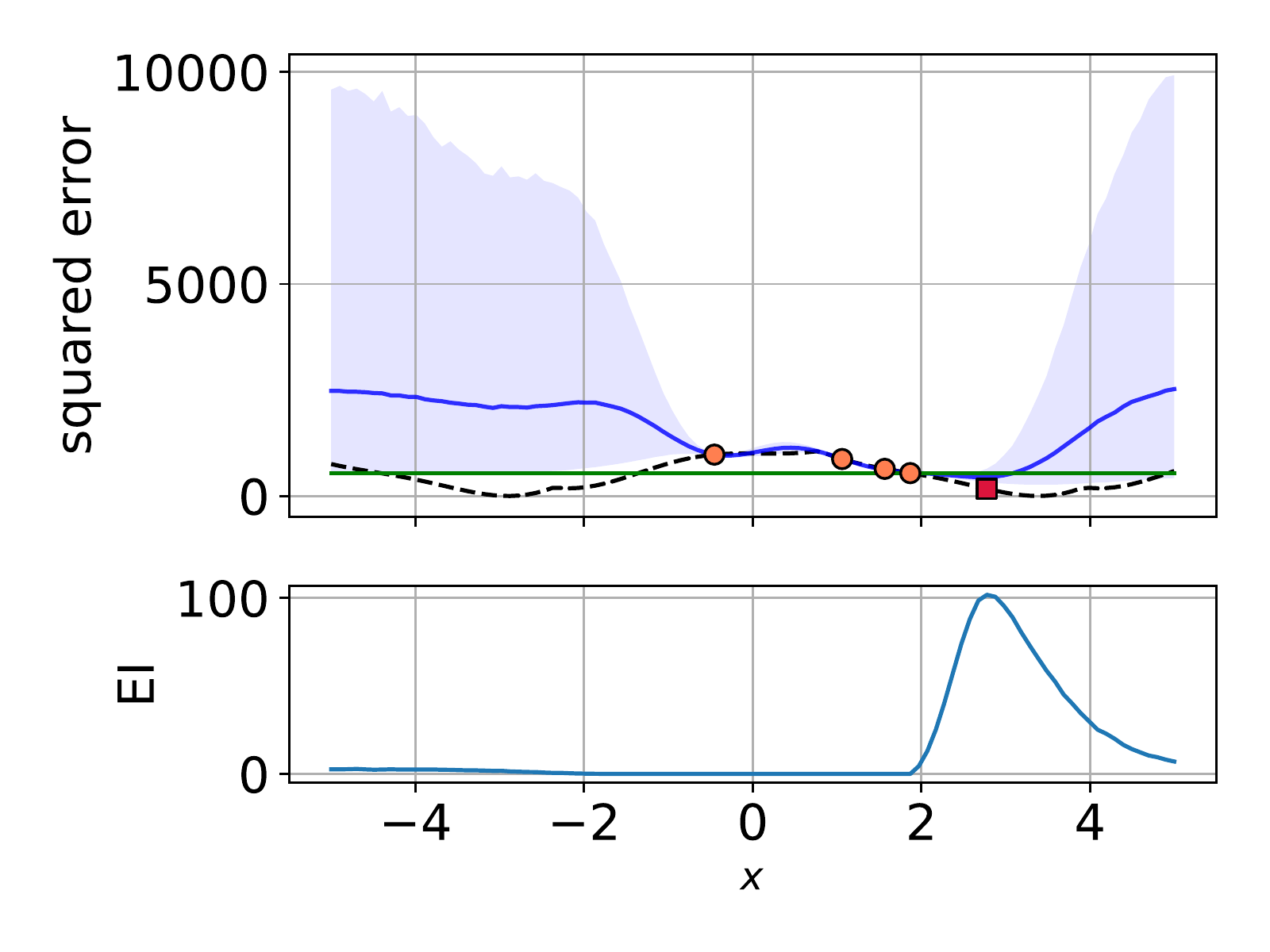}\\ 
  \hspace{-0.6cm}
  \includegraphics[bb=0 0 460 345, scale=0.32]{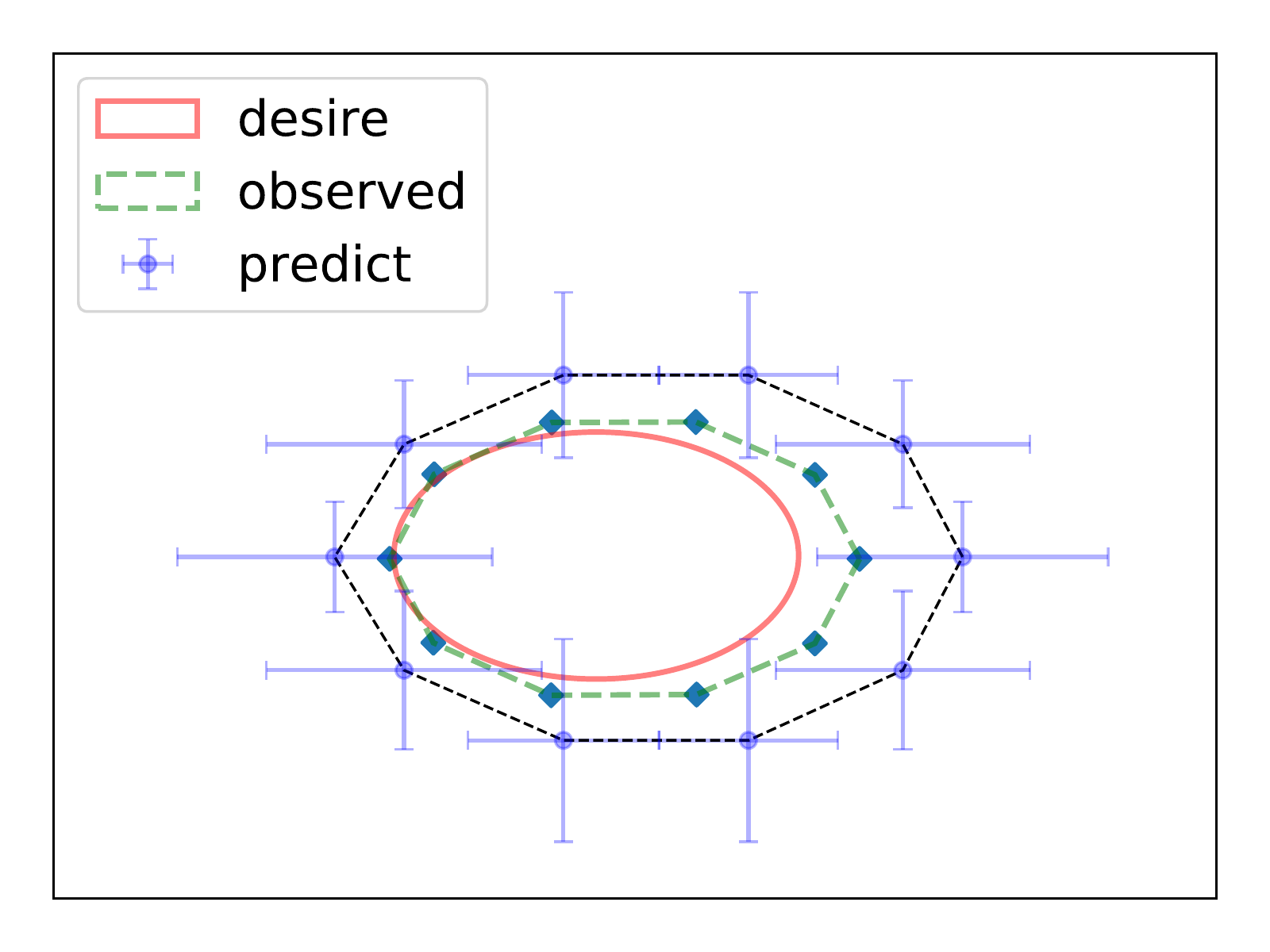} & 
  \hspace{-0.6cm}
  \includegraphics[bb=0 0 460 345, scale=0.32]{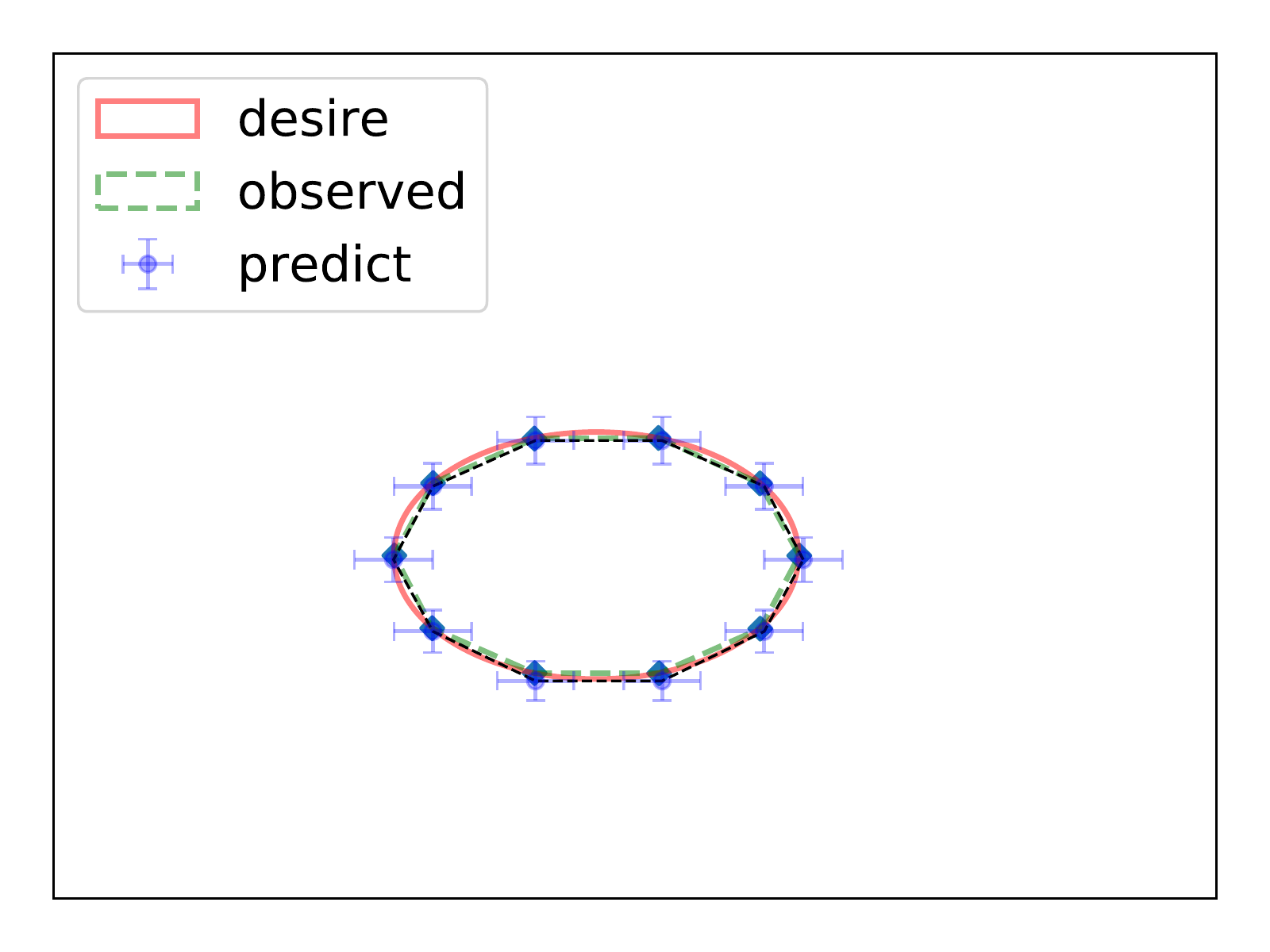} & 
  \\
  \hspace{-0.6cm}
  \includegraphics[bb=0 0 460 345, scale=0.36]{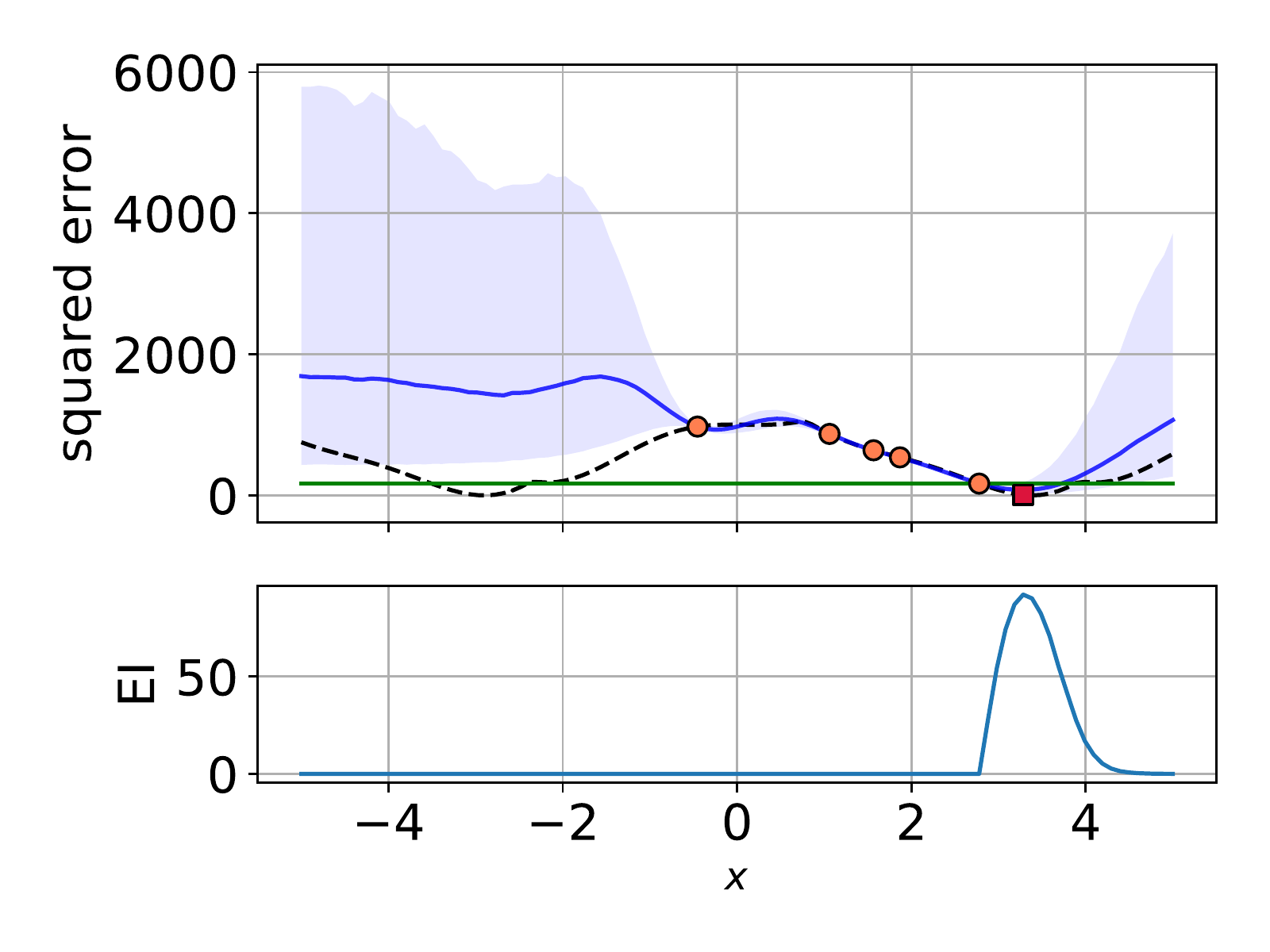} & 
  \hspace{-0.6cm}
  \includegraphics[bb=0 0 460 345, scale=0.36]{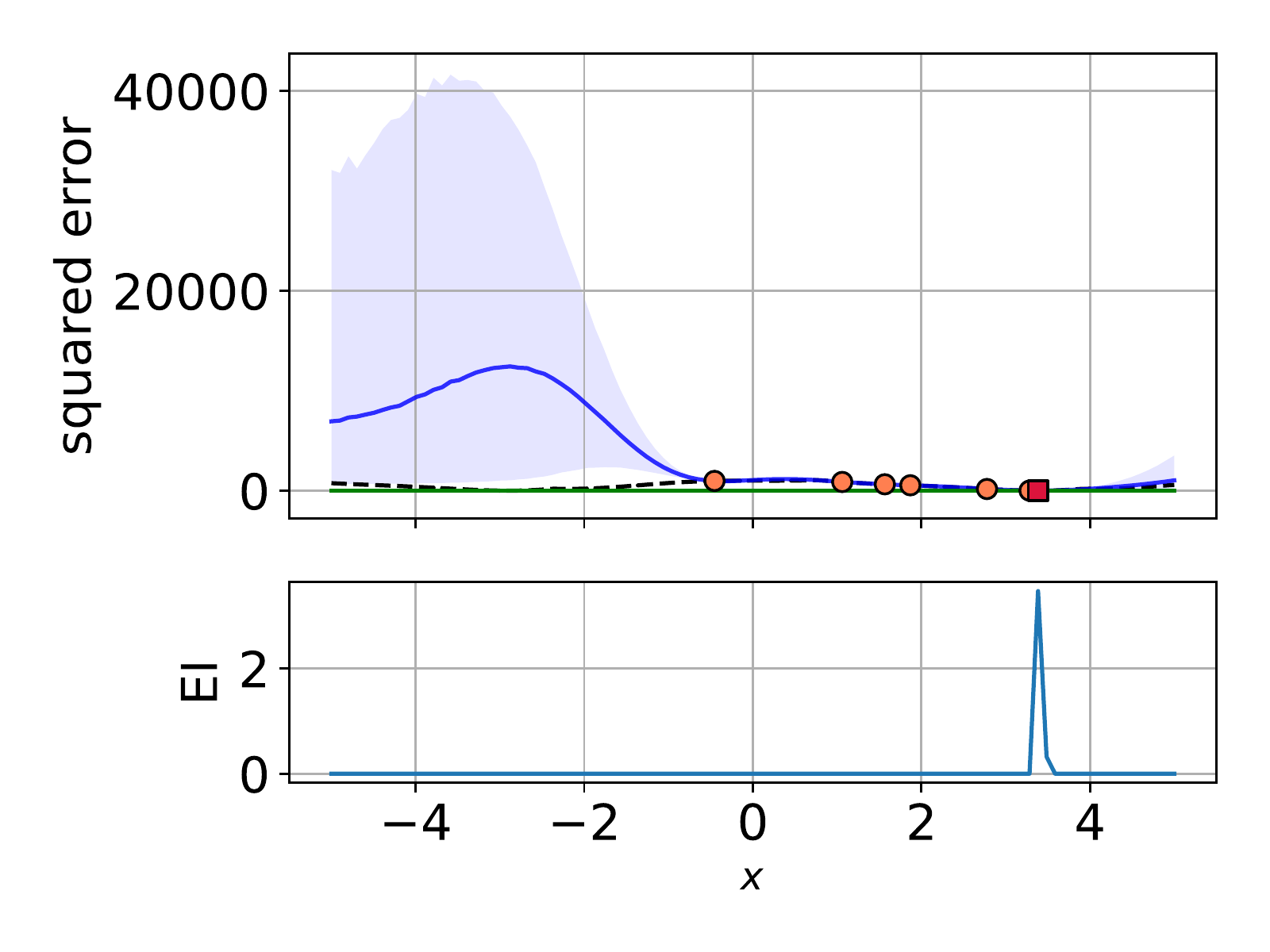}
 \end{tabular}
 \caption{
 {\footnotesize
 Results of the Sphere shape finding problem. 
 The first and third rows show the search process in the shape space, and the second and fourth rows show the models of the objective function of active learning and the EI acquisition function in the same search process. 
 In the shape plots, the colors red , blue and green represent the desired, predict and observed triangle respectively. 
 In the model plots, blue line, blue shade, black line and green line 
 represent the predictive mean, predictive variance, true squared error value and the 
 current best squared value respectively. 
 Furthermore, in the model plot, 
 the sphere points represent observed points, and square points represent the next observation point specified by EI.
 The search process proceeds from the top left to the bottom right, and the desired sphere was found in the fifth step.
 }
 \label{fig:results_circle}
 }
 \end{center}
\end{figure*}
\section{Concluding Remarks}
\label{conclusion}
We proposed a novel approach for inverse problem with structured-output. 
As a search strategy, 
two improvement-based acquisition functions, the probability of improvement and 
the expected improvement are derived for the problem where the output structure is expressed by the correlation between objective functions.
The effectiveness of the proposed method was demonstrated by experiments with artificial shape data and real material data. 
\bibliography{paper}
\bibliographystyle{abbrvnat}
\appendix

\onecolumn

\section{Implementation Details}


In the numerical experiments, we employ the Gaussian kernel as the covariance function $k$ of inputs which has two hyperparameters, 
variance and length-scale. In addition, We construct the correlation matrix $\B$ of the objective function as 
\begin{align*}
    \B = \L\L^{\top} + \kappa \I
\end{align*}
where $\L$ and $\kappa$ are hyperparameters. 
All hyperparameters are estimated via marginal likelihood 
maximization in each iteration. 
We used Python 3.7.4 and the GPy package was used for Gaussian process modeling.

\section{Details of Synthetic Data Experiments}

\subsection{The Oracle Functions}
\label{oracle_detail}
The oracle functions in the synthetic experiments are as follows:

\textbf{Triangle Data}

For the triangle shape finding problem, we prepare the 12 oracle functions. 
6 correspond to the vertices of the triangle which are given as 
\begin{align*}
    f_1(x) &= 5 \sin(x),~ 
    g_1(x) = 5 \cos(x),  \\ 
    f_2(x) &= 5 \sin(x) - \sqrt{|x|},~  
    g_2(x) = 5 \cos(x) - 2 \sqrt{|x|} \\ 
    f_3(x) &= 5 \sin(x) + \sqrt{|x|},~ 
    g_3(x) = g_2(x). 
\end{align*}
The other 6 correspond to the midpoint of each edge of the triangle and 
are given by
\begin{align*}
    f_4(x) &= 5 \sin(x) - \frac{1}{2} \sqrt{|x|},~ 
    g_4(x) = 5 \cos(x) - \sqrt{|x|} \\ 
    f_5(x) &= 5 \sin(x) + \frac{1}{2} \sqrt{|x|},~ 
    g_5(x) = g_4(x)  \\ 
    f_6(x) &= f_1(x),~ 
    g_6(x) = 5 \cos(x) - 2 \sqrt{|x|}. 
\end{align*}

\textbf{Sphere Data}

For the sphere shape finding problem, we prepare the 20 oracle functions correspond to the points on the one dimensional sphere. For each $m = 1, ..., M$ 
the oracle function is defined as 
\begin{align*}
    f_m(x) = (c(x))_0 + r(x) \cos \left(\frac{2m\pi}{M} \right),~
    g_m(x) = (c(x))_1 + r(x) \sin \left(\frac{2m\pi}{M} \right), 
\end{align*}
where $c(x)$ and $r(x)$ correspond to the center and the radius of the sphere
respectively that are given by 
\begin{align*}
    c(x) = \left( 
            \begin{array}{c}
            5 \sin(x) \\ 
            5 \cos(x)
            \end{array}
            \right),~ 
    r(x) = 5 |\sin(x) - \cos(x)|.
\end{align*}

\subsection{Experimental Results of Triangle Shape Finding Problem}
\label{triangle_results}

\begin{figure*}[t]
 \begin{center}
 \begin{tabular}{cc}
 \hspace{-0.8cm}
  \includegraphics[bb=0 0 719 539, scale=0.32]{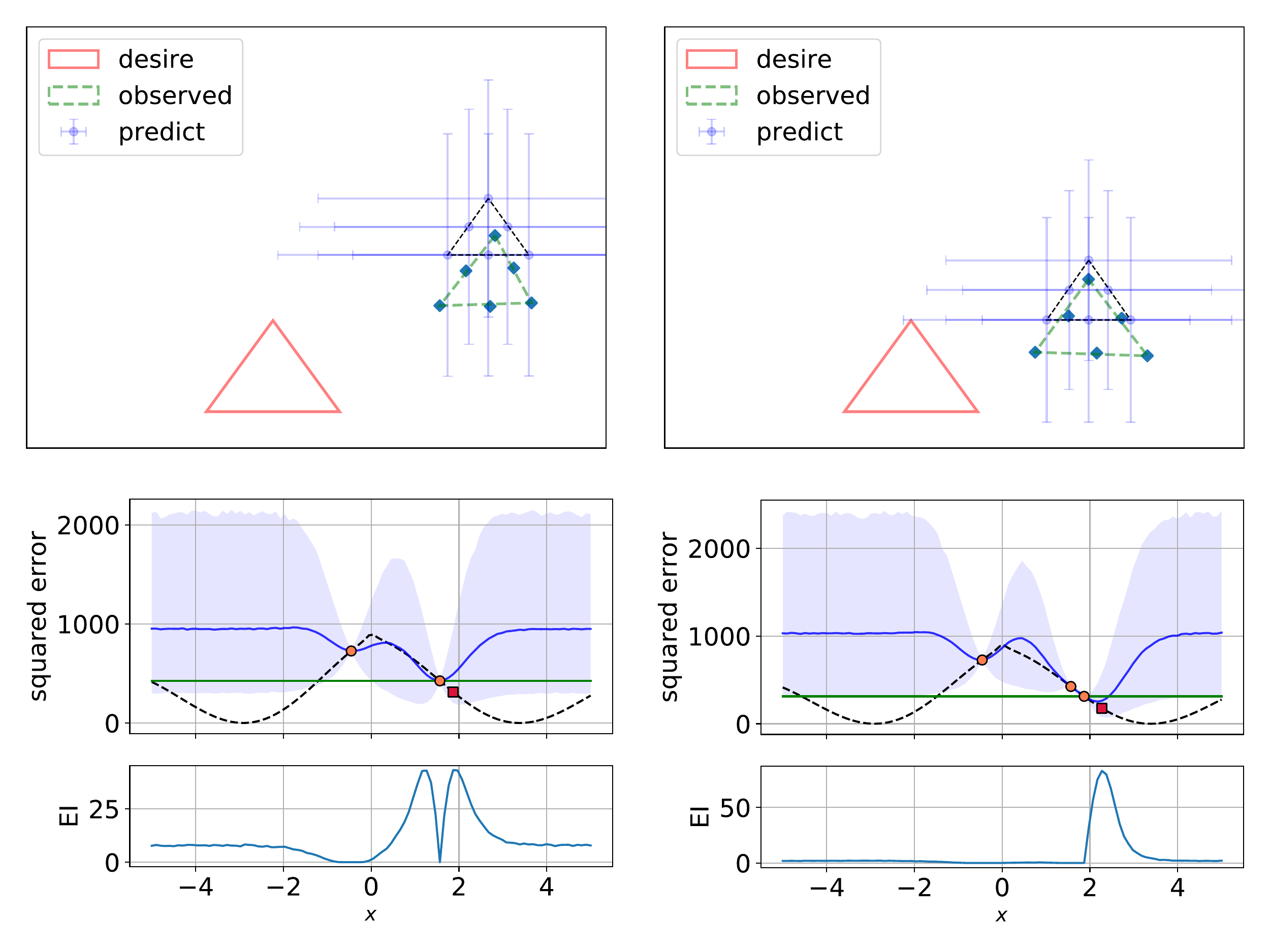} &
  \hspace{-0.6cm}
  \includegraphics[bb=0 0 719 539, scale=0.32]{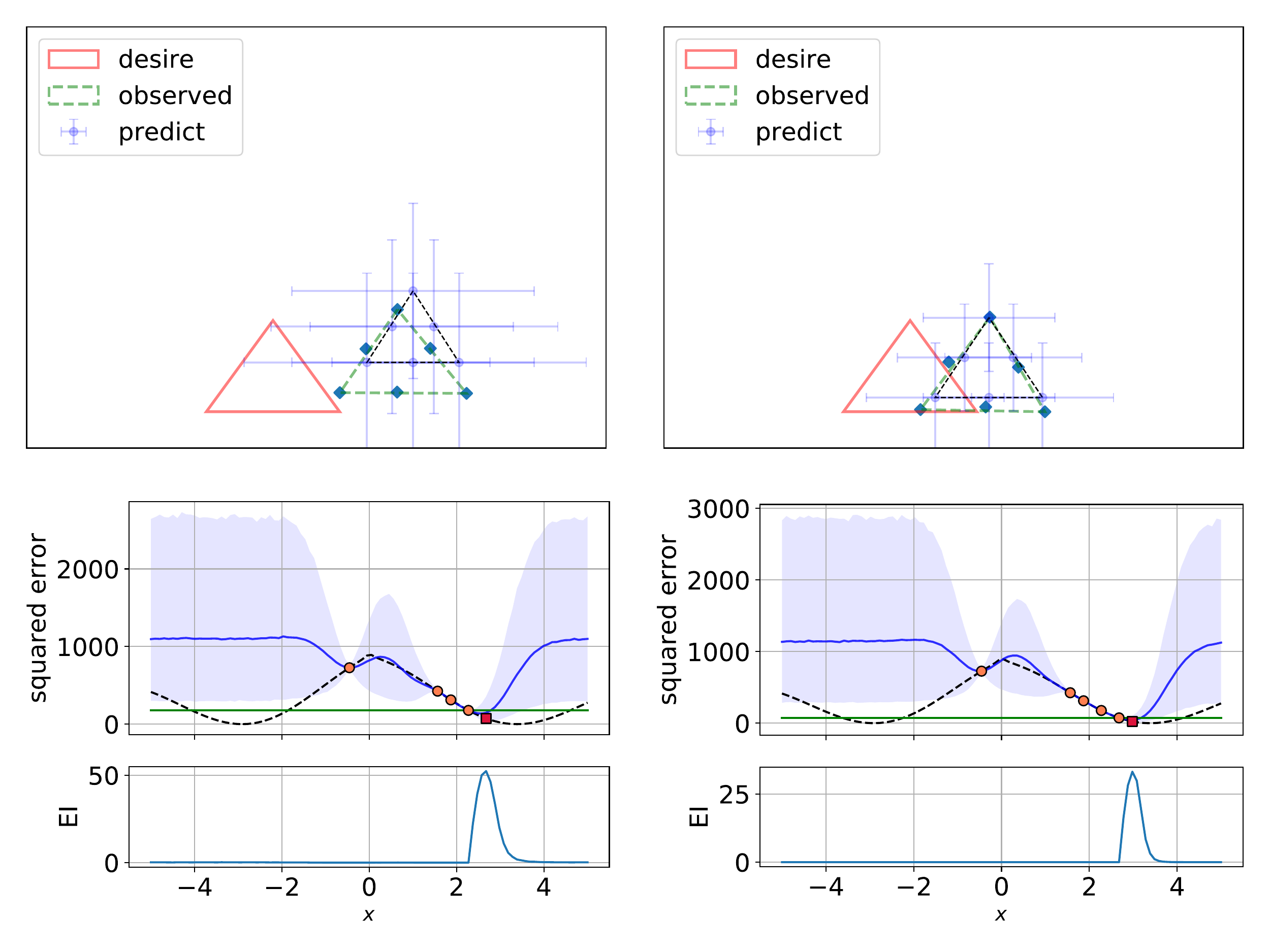} \\
  \hspace{-0.8cm}
  \includegraphics[bb=0 0 719 539, scale=0.32]{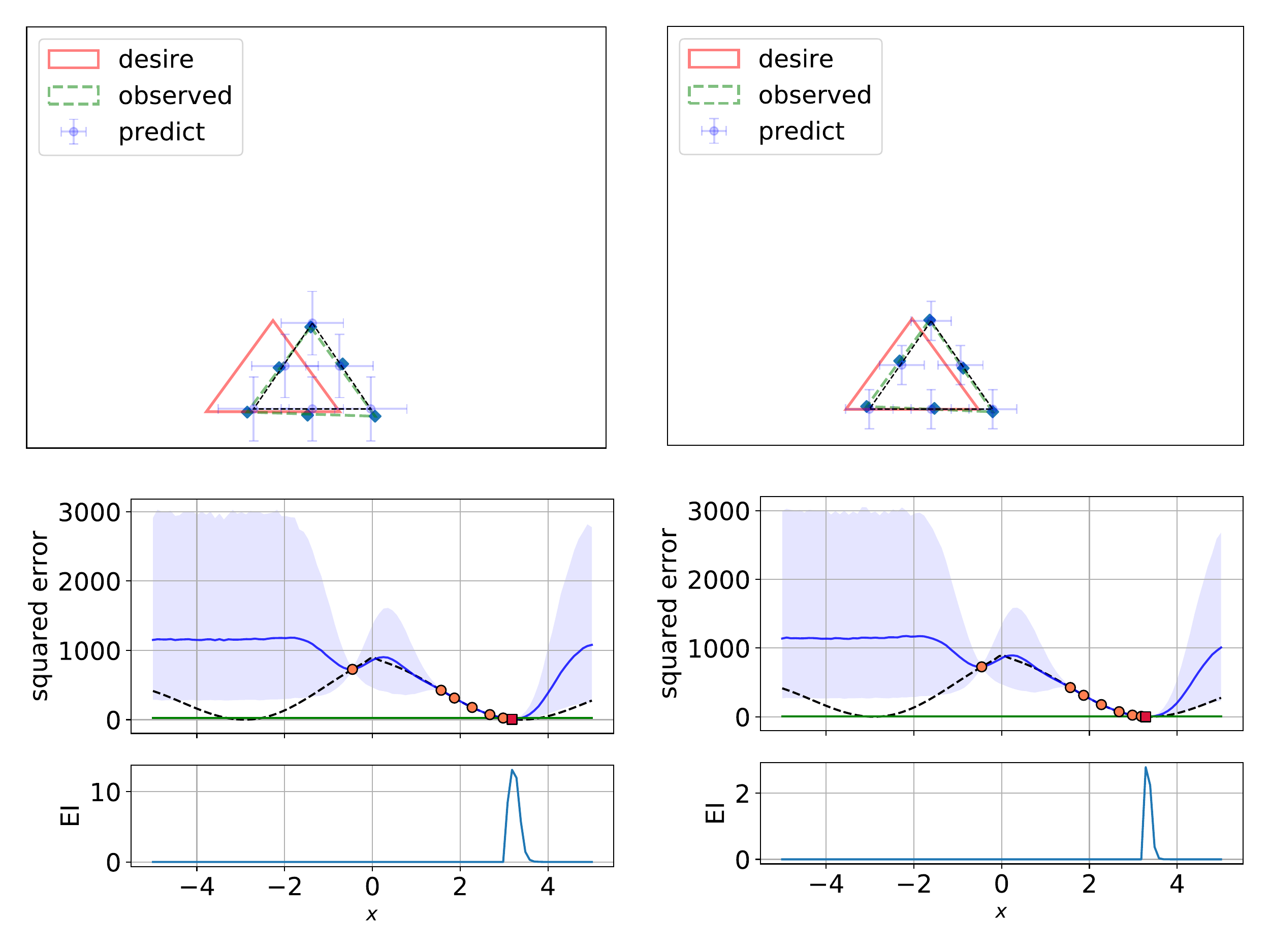} & 
  \hspace{-0.6cm}
  \includegraphics[bb=0 0 719 539, scale=0.32]{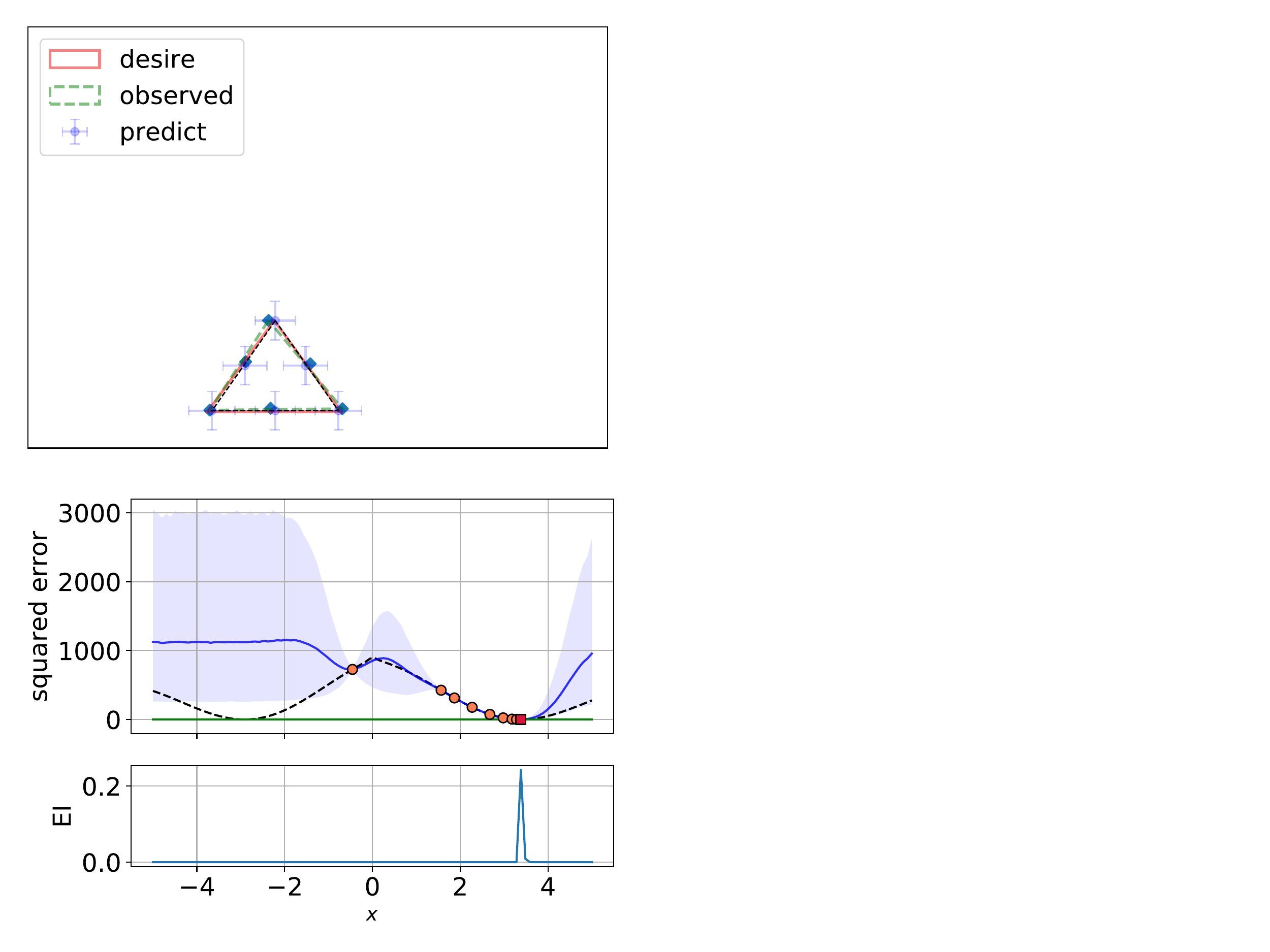} 
 \end{tabular}
  \vspace{-0.5cm}
 \caption{
 {\footnotesize
 Results of the triangle shape finding problem. 
 First and third rows show the search process in the shape space, 
 and Second and fourth rows show the models of the objective function of active learning and the EI acquisition function in the same search process. 
 In the shape plots, the colors red , blue and green represent the desired, predict and observed triangle respectively. 
 In the model plots, blue line, blue shade, black line and green line 
 represent the predictive mean, predictive variance, true squared error value and the current best squared value respectively. 
 Furthermore, in the model plot, 
 the circle points represent observed points, and square points represent the next observation point specified by EI.
 The search process proceeds from the top left to the bottom right, and the desired triangle was found in the seventh step.
 }
 \label{fig:results_triangle}
 }
 \end{center}
\end{figure*}

In Figure~\ref{fig:results_triangle}, We show the experimental results of triangle shape finding problem as in Figure~\ref{fig:results_circle}. 
As explained in Section~\ref{experiments}, the desired triangle can be 
found efficiently by the proposed method.

\end{document}